\documentclass{article}

\PassOptionsToPackage{numbers, compress}{natbib}
 \usepackage[preprint]{neurips_2025}

\usepackage[utf8]{inputenc} 
\usepackage[T1]{fontenc}    
\usepackage{hyperref}       
\usepackage{url}            
\usepackage{booktabs}       
\usepackage{amsfonts}       
\usepackage{nicefrac}       
\usepackage{microtype}      
\usepackage{xcolor}         

\usepackage{multirow}
\usepackage{multicol}
\usepackage{enumitem}
\usepackage{float}
\usepackage{graphicx}
\usepackage{subfigure}
\usepackage{array}
\usepackage{longtable}
\usepackage{amsmath}
\usepackage{amssymb}
\usepackage{mathtools}
\usepackage{amsthm}
\usepackage[most]{tcolorbox}
\theoremstyle{plain}

\theoremstyle{definition}

\theoremstyle{remark}

\theoremstyle{plain}
\newtheorem{position}{Position}[section]

\title{LLM Cannot Discover Causality, and \\Should Be Restricted to Non-Decisional Support \\in Causal Discovery}

\author{%
  Xingyu Wu\\
  The Hong Kong Polytechnic University\\
  Hong Kong SAR, China \\
  \texttt{xingy.wu@polyu.edu.hk} \\
  \And
  Kui Yu\\
  Hefei University of Technology\\
  Hefei, China\\
  \texttt{yukui@hfut.edu.cn}\\
  \And
  Jibin Wu\\
  The Hong Kong Polytechnic University\\
  Hong Kong SAR, China \\
  \texttt{jibin.wu@polyu.edu.hk} \\
  \And
  Kay Chen Tan\\
  The Hong Kong Polytechnic University\\
  Hong Kong SAR, China \\
  \texttt{kctan@polyu.edu.hk} \\
}

\begin{document}

\maketitle

\begin{abstract}
  This paper critically re-evaluates LLMs’ role in causal discovery and argues against their direct involvement in determining causal relationships. We demonstrate that LLMs’ autoregressive, correlation-driven modeling inherently lacks the theoretical grounding for causal reasoning and introduces unreliability when used as priors in causal discovery algorithms. Through empirical studies, we expose the limitations of existing LLM-based methods and reveal that deliberate prompt engineering (e.g., injecting ground-truth knowledge) could overstate their performance, helping to explain the consistently favorable results reported in much of the current literature. Based on these findings, we strictly confined LLMs’ role to a non-decisional auxiliary capacity: \emph{LLMs should not participate in determining the existence or directionality of causal relationships, but can assist the search process for causal graphs} (e.g., LLM-based heuristic search). Experiments across various settings confirm that, by strictly isolating LLMs from causal decision-making, LLM-guided heuristic search can accelerate the convergence and outperform both traditional and LLM-based methods in causal structure learning. We conclude with a call for the community to shift focus from naively applying LLMs to developing specialized models and training method that respect the core principles of causal discovery.
\end{abstract}

\section{Introduction}

Causal discovery aims to uncover underlying causal relationships from observational data \cite{glymour2019review}. Traditional causal discovery algorithms (CDAs) construct causal structures by systematically leveraging conditional independence tests, optimization techniques, or functional causal models. These methods integrate theoretical principles from graphical models, information theory, and structural causal modeling to ensure interpretability and maintain theoretical guarantees in causal discovery \cite{zanga2022survey}. Recently, large language models (LLMs) have emerged as a potential alternative for causal discovery \cite{kiciman2023causal}. Many studies have explored the application of LLMs in this domain, leveraging their linguistic knowledge and common-sense reasoning to identify causal relationships. These efforts have sparked significant interest, particularly in investigating whether LLMs can assist in identifying causality from text and structured data.

LLMs have achieved notable performance in causal discovery, particularly in identifying pairwise causal relationships from text \cite{jincan,ban2023query}. Existing studies have leveraged variable information \cite{constantinou2024using} and background knowledge \cite{zhang2023understanding} to enhance LLMs’ ability to identify causality. These studies not only highlight LLMs’ strengths in recognizing pairwise causality but also explore hybrid approaches that integrate LLMs with traditional CDAs to enhance performance \cite{long2023causal}. For instance, researchers have utilized LLMs to initialize causal structures \cite{naik2024applying} or refine CDA-generated graphs \cite{khatibi2024alcm}. Most studies demonstrate promising results, suggesting that LLMs hold significant potential in causal discovery. However, these efforts primarily focus on empirical evaluations, leaving open questions about whether LLMs truly possess causal reasoning abilities. Some researchers have raised concerns that LLMs may struggle with complex causal tasks \cite{zhang2023understanding} and are prone to generating hallucinated causal relations, i.e., assigning causality where none exists \cite{zevceviccausal}. Given the instability of LLMs, several mitigation strategies have been proposed. These include providing observational data \cite{cai2023knowledge}, employing LLMs as conditional independence test operators \cite{zhang2024causal}, and incorporating LLM-generated causal judgments into classical scoring functions as priors \cite{ban2023query}. While these improvements claim to enhance reliability to some extent, they do not fully resolve the fundamental limitations of LLMs in causal discovery.

\begin{figure}[t]
  \centering
  \includegraphics[width=13.8cm]{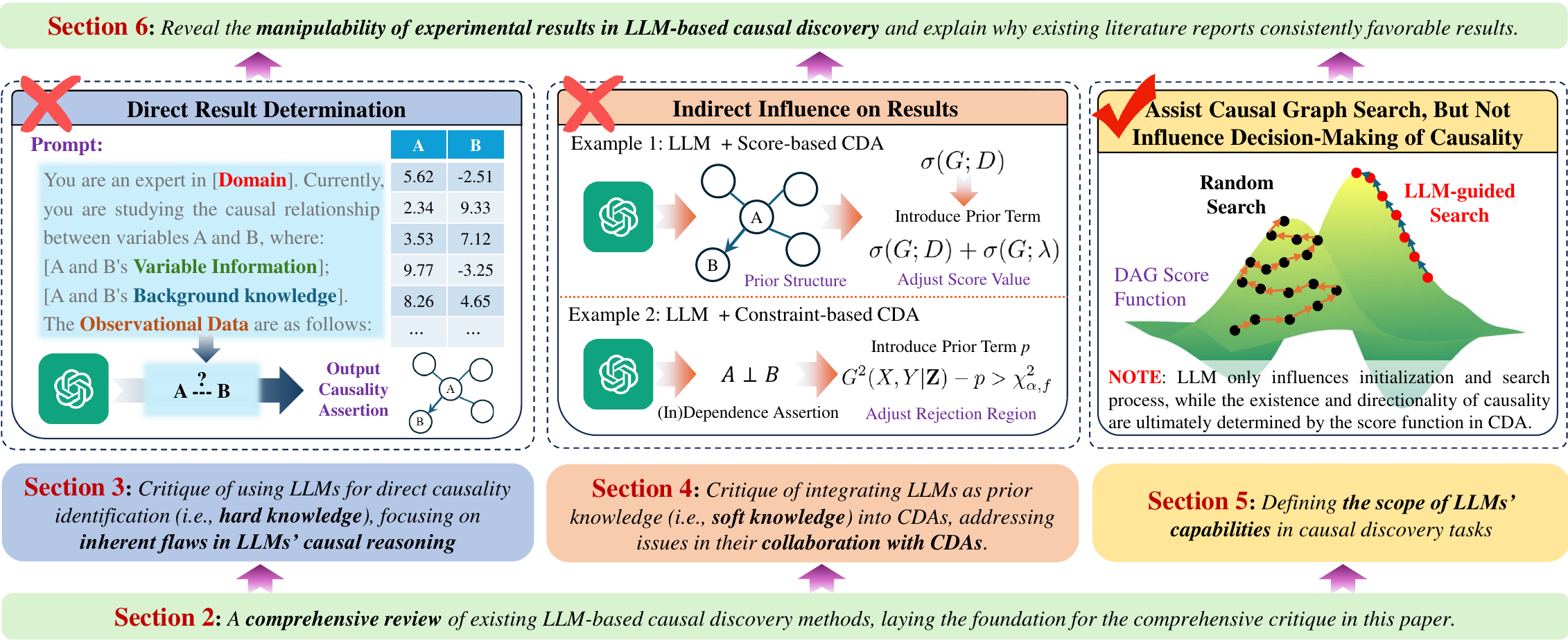}
  \caption{This logical framework illustrates the focus and main positions of Sections 2 to 6. Building on Section 2’s comprehensive review of existing methods, Sections 3 and 4 conduct a thorough critique of current algorithms, whether using LLMs as hard knowledge or soft knowledge. Then, Section 5 defines the scope of LLMs’ role in causal discovery. Section 6 further reveals the manipulability of experimental results in existing methods, which explains why they achieve good performance.} \label{introfig}
\end{figure}

Different from most existing studies, we argue that the role of LLMs in causal discovery tasks is limited. Through the following critical and compensatory perspectives, we clearly define the current scope of LLMs' applicable functions: 
\begin{enumerate}[leftmargin=2em]
\item \textbf{Critical Position: LLMs cannot identify causality, and their outputs should not directly or indirectly determine the existence or directionality of causal relationships.} This includes:
    \begin{itemize}
      \item \textbf{Direct Result Determination}: LLMs should not independently conclude whether a causal relationship exists (\underline{\textbf{Section \ref{sec3}}}).  
      \item \textbf{Indirect Influence on Results}: Causal relationships identified by LLMs should not be embedded as "prior knowledge", whether as hard constraints or soft penalties, into the decision-making processes of CDAs (\underline{\textbf{Section \ref{sec4}}}).  
    \end{itemize}
\item \textbf{Compensatory Position: Without influencing causal decision-making, LLMs can assist CDAs by improving the search process for causal graphs, thereby accelerating the convergence toward optimal causal structures (\underline{Section \ref{sec5}}).}
\end{enumerate}

To this end, this paper investigates the limitations of LLMs in causal discovery from two critical perspectives: (i) the intrinsic limitations of LLMs to infer causality, and (ii) the challenges arising from their integration with classical CDAs. As shown in Figure \ref{introfig}, Section \ref{sec3} first explains how LLMs' autoregressive modeling based on word order \cite{zhao2023survey} inherently differs from the probability decomposition in structural causal models \cite{bongers2021foundations}, leading to a lack of theoretical grounding for causal reasoning. Through a series of empirical studies, we systematically analyze the impact of several key factors—such as word order, redundancy, entity positioning, and numerical precision—on LLMs’ causal inference. Results reveal that LLM performance is highly sensitive to prompt design and text quality, and merely supplying observational data does not guarantee improved reliability. Unless causal relationships are stated with exceptional clarity and minimal noise, LLMs often fail to reach consistent or correct conclusions. Given these reliability concerns, Section \ref{sec4} further investigates the interaction between LLMs and traditional CDAs, analyzing different collaboration paradigms and their limitations. We focus on how using LLM-derived causal relationships as (hard/soft) prior knowledge in CDAs \cite{ban2023query} compromises algorithmic reliability and theoretical guarantees, regardless of whether the method is score-based or constraint-based. Moreover, in Section \ref{sec6}, we demonstrate that the performance of LLM-based methods can be artificially inflated via prompt engineering, for instance, by injecting ground-truth into the input.

Building on these analyses, we advocate for a principled delineation of LLMs’ functional boundaries in causal discovery: LLMs should not determine the existence or directionality of causal relationships, but can serve as auxiliary agents that guide the causal graph search process. Section \ref{sec5} introduces a novel paradigm where LLMs assist in heuristic search—such as through initialization, mutation guidance, or cycle resolution—while preserving the decision-making authority of theoretically sound CDAs. By leveraging LLMs’ general world knowledge \cite{wu2024evolutionary} in a non-decisional role, this framework helps CDAs escape local optima and converge faster to globally optimal structures. Crucially, the evaluation of causal structures remains grounded in statistical analysis of observational data via established scoring functions or independence tests, which is fundamentally different from existing LLM-based causal discovery methods. Comprehensive experiments across varying network scales confirm that when LLMs are appropriately confined to assistive roles, they can significantly enhance search efficiency and accuracy compared to both traditional CDAs and existing LLM-based methods. Thus, our results underscore the value of LLMs as heuristic facilitators—not causal reasoners—in the pursuit of reliable causal discovery.

In summary, this paper presents a critical re-evaluation of the role of LLMs in causal discovery, challenging the prevailing optimism surrounding their application. We call upon the research community to adopt a more principled and cautious stance when incorporating LLMs into causal tasks, ensuring that empirical convenience does not come at the cost of epistemic soundness. The path forward lies not in overstating the reasoning capabilities of unmodified general-purpose LLMs; instead, the future of LLM-based causal discovery depends on rethinking model design and training objectives to align with the structural and statistical principles of causality.

\section{Characteristics of LLM-based Causal Discovery (Detailed in Appendix \ref{app_summary})}
\label{sec2}

Causal discovery, a fundamental task in causal inference \cite{glymour2019review}, aims to uncover causal relationships from observational data, with traditional methods like constraint-based and score-based approaches \cite{vowels2022d} relying on statistical and mathematical frameworks. Recent advancements in LLMs have introduced new opportunities, as LLM-based methods leverage linguistic knowledge, common sense, and domain insights to complement traditional techniques. Existing approaches, summarized in Table \ref{tab_summary} in \textbf{Appendix} \ref{app_summary}, can be categorized along three dimensions: (1) \textbf{Pairwise Causality and Global Causal Graph}: Methods focus on either inferring pairwise causal directions/existence \cite{kiciman2023causal,zevceviccausal,jincan,chiunveiling} or learning full causal graphs by integrating LLMs with data-driven CDAs \cite{long2023causal,ban2023causal}; (2) \textbf{Prompt Design}: Strategies include incorporating variable information \cite{constantinou2024using}, background knowledge \cite{zhang2023understanding,zhang2024causal}, observational data \cite{cai2023knowledge}, and CDA results \cite{abdulaal2023causal} to guide LLMs; (3) \textbf{Collaboration Paradigms}: Hybrid approaches involve using LLMs for initialization/prior \cite{ban2023query,darvariu2024large,naik2024applying}, post-processing \cite{long2023causal,castelnovo2023marrying,khatibi2024alcm}, iterative collaboration \cite{abdulaal2023causal,ban2023causal,takayama2024integrating}, or modular integration \cite{zhang2024causal,cohrs2024large} with traditional CDAs. In \textbf{Appendix} \ref{app_summary}, we detail the comprehensive overview of existing LLM-based causal discovery methods.

\section{Fundamental Limitations of LLMs in Identifying Causality}
\label{sec3}

\subsection{Identify Causality in Textual Data}

We commence by elucidating that the autoregressive paradigm of LLMs, which is predicated on word order \cite{zhao2023survey}, is not conductive for discerning causality. LLMs model the associations between words through the attention mechanism. Fundamentally, this amounts to modeling the conditional probability contingent on the preceding context. Consider an input word sequence \(\mathbf{x}=(x_1,x_2,\ldots,x_T)\) fed into the LLM, where \(T\) denotes the sequence length. The attention mechanism empowers the model, when generating or predicting each word \(x_i\), to incorporate the information from all prior words \(x_1,x_2,\ldots,x_{i - 1}\) (or words within a specific window). Consequently, the LLM is effectively modeling the following distribution:
\begin{equation}\label{eq_1}
  P(\mathbf{x}) = P(x_1)\cdot P(x_2\mid x_1)\cdots P(x_T\mid x_1,x_2,\ldots,x_{T - 1})
\end{equation}
By the chain rule, Eq. (\ref{eq_1}) holds true for any sequence \(\mathbf{x}\). From the perspective of probability decomposition, this word order-based decomposition is significantly different from the decomposition based on causal graphs in structural causal models \cite{bongers2021foundations}. For a causal structure represented by a directed acyclic graph (DAG), the structural causal model decomposes its probability distribution as:
\begin{equation}\label{eq_2}
P(X_1,X_2,\ldots,X_n)=\prod_{i = 1}^{n}P(X_i|pa(X_i))
\end{equation}
where \(pa(X_i)\) is the set of parent nodes of \(X_i\). This decomposition method in Eq. (\ref{eq_2}) can accurately reflect the conditional dependence and independence relationships between variables, which cannot be captured by the word order-based modeling paradigm in Eq. (\ref{eq_1}), because when determining the generation distribution at a certain position, all previous words are considered. For example, in a text, there may be a causal relationship: ``Because it rained (\(X\)), the ground is wet (\(Z\)), causing people to walk carefully (\(Y\))". When the LLM generates the probability of \(Y\), it will consider the word \(X\), even though in the causal structure, given \(Z\), \(X\) and \(Y\) are conditionally independent (causal chain \(X\rightarrow Z\rightarrow Y\)). On the other hand, the sequential order of words may lead to confusion in parsing causal relationships. For example, in a text, there are two events \(A\) and \(B\) that are causally independent, but \(A\) comes before \(B\) in the text order. The LLM will be affected by \(A\) when estimating \(P(B)\), rather than estimating according to their true causal relationship.

Therefore, the modeling approach of LLMs is not oriented towards causal relationships, but rather more towards modeling the correlations between words in the text. At the same time, the variable relationships constructed by LLMs contain a large amount of redundant structure, which is essentially different from the sparse characteristics of causal graphs. From this perspective, there is no theoretical basis for LLMs to identify causal relationships, and the causal relationships obtained by LLMs from texts are not robust, as shown in the following position. 
\begin{position}\label{pos_1}
  When LLMs identify causal relationships from text, they are at least affected by the following factors: (1) The word order in the text and whether it follows the order of causal words that frequently appear in the LLM's training corpus; (2) The amount of redundant entities and redundant information in the text; (3) The distance between the entities whose causal relationships are to be analyzed in the text. 
\end{position}

\textbf{Factor Analyses and Empirical Studies}: As shown in Position \ref{pos_1}, LLMs’ ability to infer causal relationships is influenced by multiple textual factors, including word order, the presence of redundant entities, and the textual distance between causally related entities. To systematically investigate these limitations, we conducted a series of empirical analyses examining the impact of these factors on mainstream LLMs' causal identification performance. The experimental results, along with detailed discussions on how these factors affect LLMs' reasoning processes, are presented in \textbf{Appendix} \ref{app_exsec3_1}.

\subsection{Identify Causality in Observational Data}

The causal relationships inferred by LLMs from textual information are often influenced by subjectivity, ambiguity, and incompleteness inherent in textual data. In response, recent studies have explored using LLMs to analyze observational data directly for causal discovery \cite{cai2023knowledge}. Unlike textual information, observational data provides an objective and detailed record of real-world phenomena, offering richer and more accurate information. By allowing LLMs to analyze observational data, it is possible to compensate for the limitations of textual sources, enabling the models to explore causal relationships from multiple dimensions. Researchers aim for LLMs to infer causality more effectively by examining correlations and covariances among variables in the data. However, we observe that even when supplied with observational data, the architecture and operational principles of LLMs may still limit their capacity to understand and accurately infer causal relationships. Unlike traditional CDAs \cite{niu2024comprehensive}, which directly manipulate numerical data through arithmetic operations (e.g., addition and multiplication), LLMs process data as tokenized strings \cite{rajaraman2024toward}. Numerical data encoded as tokens can distort original numerical features, hindering LLMs' ability to grasp true causal mechanisms. This highlights inherent limitations in LLMs when processing numerical data for causal discovery, as proposed in the following Position:
\begin{position}\label{pos_2}
  Unlike traditional CDAs, LLMs lack intrinsic mechanisms to exploit numerical features. LLMs fail to effectively utilize high-precision numerical data for inferring causal patterns. Even with low-precision data, their causal discovery capabilities are constrained.
\end{position}

\textbf{Empirical Studies}: To systematically evaluate these limitations, we conduct a series of controlled experiments assessing the sensitivity of LLMs to numerical precision, their ability to recognize correlations, and their performance on benchmark causal discovery tasks. The detailed experimental setup, results, and further discussions on these limitations are presented in \textbf{Appendix}~\ref{app_exsec3_2}.

\section{Risks of Integrating LLMs with CDAs}
\label{sec4}

Since the causal relationships derived from LLMs in textual and observational data are unreliable, recent studies have begun to explore the integration of LLMs with traditional CDAs \cite{ban2025llm,long2023causal}. This section will discuss the issues that arise in the interaction between LLMs and these CDAs. In Section \ref{sec2}, we discussed four ways in which LLMs can be integrated with CDAs. The second method, which treats the LLM as a posterior, essentially places the decision-making process under the control of the LLM. This approach does not involve mutual correction or guidance between the CDA and the LLM; instead, it narrows the decision space to some extent (e.g., by using equivalence classes derived from the algorithm). The CDA relies on the LLM to resolve issues that it cannot handle, but fundamentally, it places full trust in the LLM’s decisions in this space and does not involve a collaboration between the algorithm and the LLM. The issues faced by this approach are the same as those discussed in the previous section, and therefore, we will not explore them further in this section.

\textbf{Research of Reliable Prior Knowledge}: Other than Type (2), other types mentioned in Section \ref{sec2} all involve using the results from LLMs as priors for CDAs. This is currently the mainstream paradigm for combining LLMs with traditional CDAs. In fact, the use of prior knowledge in causal discovery was studied even before the advent of LLMs. However, these prior knowledge studies typically assume reliable background knowledge \cite{meek1995causal,andrews2020completeness,wang2023sound,wang2024new}. In causal discovery research, there is a significant difference between reliable and unreliable prior knowledge. Reliable prior knowledge is typically based on rigorous research, theories, or empirical evidence, and it is characterized by a high degree of certainty and accuracy. For example, background knowledge defined in \cite{meek1995causal} specifies directed edges that are either prohibited or required, which can directly guide the construction and orientation of the graph in an algorithm. However, as discussed in Section \ref{sec3}, LLMs clearly fall under the category of unreliable prior knowledge. Due to the quality of textual data and model limitations, LLMs contain numerous errors and uncertainties, making it difficult to accurately assess their reliability. 

Reliable prior knowledge is strongly consistent with causal models and can effectively assist in their construction and inference. Some existing studies use reliable background knowledge to narrow the search space for causal relationships, improving inference accuracy and efficiency \cite{meek1995causal,andrews2020completeness,wang2023sound,wang2024new}. On the other hand, unreliable prior knowledge may not align with the true causal relationships. A potential risk is that if the LLM misjudges the causal relationships, the algorithm may search in the wrong direction, making it harder to find the correct causal relationships. Furthermore, the uncertainty in LLM-generated knowledge leads to instability in causal discovery results, affecting model reproducibility and making it difficult to provide stable and reliable causal inferences in practical applications. We present the following viewpoint and analyze the issues faced by constraint-based methods and score-based methods when embedding LLM priors.
\begin{position}
  When using causal relationships identified by LLMs as priors in CDAs, the unreliable prior knowledge provided by LLMs undermines the reliability and theoretical guarantees of these algorithms.
\end{position}
\textbf{Analysis of Score-based Methods}: A significant body of research \cite{ban2023query,ban2023causal,chen2023mitigating,ban2025llm} embeds LLM results as prior knowledge within classical scoring functions. Due to data limitations and the uncertainty of prior knowledge, it may not always be realistic to strictly adhere to all constraints. Therefore, soft constraint methods have been proposed. The core idea is to introduce a fault tolerance mechanism when applying constraints, allowing some flexibility to accommodate potential conflicts between data and prior knowledge. This is achieved by modifying the scoring function, incorporating the prior constraints in the form of Bayesian priors, rewarding networks that satisfy specific structural constraints, and penalizing those that do not. According to Bayes' theorem, the relationship between causal structure (\(G\)), observed data (\(D\)), and prior constraints (\(\lambda\)) can be expressed as:
\begin{equation}\label{eq_4}
  P(G | D, \lambda) = \frac{P(D | G, \lambda) P(G | \lambda)}{P(D | \lambda)}
\end{equation}

Studies such as \cite{ban2023query,ban2023causal,chen2023mitigating,ban2025llm} rewrite the causal structure scoring function as follows:
\begin{equation}\label{eq_5}
  \sigma(G; D) \rightarrow \sigma(G; D, \lambda) = \sigma(G; D) + \sigma(G; \lambda)
\end{equation}
Here, \(\sigma(G; D)\) is the scoring function used in causal structure learning, such as the BIC or BDeu score, while \(\sigma(G; \lambda)\) is computed based on the prior constraints \(\lambda\), considering the probability of specific structures.

The first problem with this type of method is the direct addition of $\sigma(G ; D)$ and $\sigma(G ; \lambda)$, while prior information and data are in different probability spaces. Specifically, $\sigma(G; D)$ reflects the log-likelihood of the data, which is calculated under the sample data $D$. $\sigma(G; \lambda)$ reflects the prior probability of the causal structure, which is independent of the data and is defined by the LLM. Here, $\lambda$ is a soft-constraint converted from the results generated by the LLM according to certain preset rules, and it is usually not a strict probability distribution. This means that the direct addition of $\sigma(G ; D)$ and $\sigma(G ; \lambda)$ lacks mathematical correctness. In addition, the scales of the two are not compatible. Without proper normalization, the scales of the data-scoring term and the prior-scoring term may be extremely mismatched, causing one term to dominate the final scoring function. Taking the BIC score as an example:
\begin{equation}\label{eq_6}
  \sigma_{\text{BIC}}(G; D)=\log P(D | G)-\frac{k}{2}\log N
\end{equation}
where $k$ is the number of free parameters of the model, and the sample size $N$ significantly affects the order of magnitude of the score value. $\sigma(G; \lambda)$ which is independent of $N$, if not properly scaled, its impact may be drowned out by the data-scoring term, or conversely, it may have an unreasonable dominant effect on the final score. Taking the MDL score as another example:
\begin{equation}\label{eq_7}
  \sigma_{\text{MDL}}(G; D)=L(D | G)+L(G)
\end{equation}
where $L(D | G)$ is the encoding length of the data given the network $G$, and $L(G)$ is the encoding length of the network structure (structural complexity penalty term). They are usually measured in bits, and may not even match the dimension of the prior term $\sigma(G; \lambda)$. Examples of non-matching dimensions also include cross-validation based \cite{huang2018generalized} and structure entropy based scoring functions \cite{de2018entropy}. Moreover, some scoring functions already have priors, and the direct introduction of $\sigma(G ; \lambda)$ conflicts with the existing priors. For example, the BDeu score by default uses the Dirichlet uniform prior and assumes prior independence of all variables \cite{suzuki2017theoretical}, which is clearly contrary to the recognition results of the LLM. There is already a hyper - parameter in BDeu to control the influence of data on the posterior. After introducing $\sigma(G ; \lambda)$, it is even more necessary to balance the influence of the two priors and the data, which cannot be solved by simply adding $\sigma(G ; \lambda)$.

In short, if $\sigma(G ; D)$ in existing studies uses classical causal network scoring functions, they do not directly calculate $\log P(D|G)$, but calculate some other quantity related to $\log P(D|G)$. These quantities are usually approximations or positively correlated with the degree of data fitting, and at the same time, they carry other priors and penalty terms. And $\sigma(G ; \lambda)$ is the probability modeled for a specific structure according to the output of the LLM, and then the corresponding score is obtained, which does not depend on the specific statistical characteristics of the data. The consequence of direct addition is that the two parts of the scoring function are superimposed under different dimensions, which may lead to an unreasonable trade - off between data fitting and prior knowledge in the model. Eventually, $\sigma(G ; D)$ and $\sigma(G ; \lambda)$ produce scores of different scales in different probability spaces, and direct summation is very likely to cause one term to have too much influence on the overall scoring function, thus distorting the final result of causal structure learning.

The second problem of this type of methods is that introducing the LLM as a prior will undermine the theoretical basis and basic properties of causal network scoring function, such as Decomposablity and Score Local Consistency \cite{chickering2002optimal}, which makes certain optimization algorithms inapplicable. In traditional causal discovery settings, a decomposable scoring function allows the overall causal structure score to be broken down into the sum of local scores for each variable and its parent set, i.e.,
\begin{equation}\label{eq_8}
  \sigma(G ; D) = \sum_{i = 1}^{n}\sigma(v_{i}, \text{pa}(v_{i}); D)
\end{equation}
This property makes it easier to compute and understand the scoring function, and also provides the foundation for many efficient causal structure learning algorithms, as these algorithms can independently evaluate and optimize each local structure. However, after introducing the soft constraint scoring function \(\sigma(G ; D, \lambda) = \sigma(G ; D) + \sigma(G ; \lambda)\), for \(\sigma(G ; \lambda)\), which is computed based on the prior constraints \(\lambda\), the calculation sometimes depends on global features of the entire causal structure \(G\) rather than being naturally decomposable into local variables and their parent sets as in \(\sigma(G ; D)\). For example, prior constraints may impose complex relationships among multiple variables, which cannot be easily decomposed into individual constraints for each variable. This leads to the inapplicability of many optimization algorithms designed based on decomposability. 

On the other hand, the introduction of prior constraints \(\sigma(G ; \lambda)\) in the soft constraint scoring function disrupts Score Local Consistency. Prior knowledge is often based on external assumptions or empirical observations, which may not always be accurate or applicable. When the local structure of the causal graph changes, the changes in \(\sigma(G ; \lambda)\) may not align with the actual relationships between variables. For example, prior knowledge may erroneously emphasize certain causal relationships between variables, and when the local structure changes to better reflect the data characteristics, the prior constraints may still assign high scores to the incorrect structure, or give unreasonably low scores to the newly proposed structure. This undermines the ability of the soft constraint scoring function to accurately reflect the validity of local structural changes. These issues break the theoretical foundations of the scoring function in practical applications.

\textbf{Analysis of Constraint-Based Methods}: The core of constraint-based methods is conditional independence testing. The simplest way to embed LLMs is through textual prompts, asking the LLM to determine conditional dependencies and independencies between variables \cite{zhang2024causal, cohrs2024large}, which corresponds to the fourth method discussed in Section \ref{sec2}. The main issues with this approach have already been analyzed in Section \ref{sec3}. Another way to incorporate unreliable priors is by considering prior information in the computation of the statistical measures for conditional independence tests. Some studies have made related attempts in causal feature selection, a downstream task in causal discovery. For discrete variables, constraint-based methods typically use the \(G^2\)-test to determine the conditional dependence and independence between variables. Research such as \cite{pmlr-v22-pocock12, wu2022domain, wu2023feature} constructs different prior terms \(p\) based on background knowledge and uses \(p\) to adjust the computation of the \(G^2\)-statistic \cite{pearl2009causality}, embedding the prior into the hypothesis testing process. The expression is:
\begin{equation}\label{eq_9}
  G^2(X,Y|\mathbf{Z}) - p > \chi^{2}_{\alpha,f}
\end{equation}
where \(p\) is calculated based on prior knowledge, \(\alpha\) is the significance level of the hypothesis test, and \(f\) is the degrees of freedom of the \(G^2(X,Y|\mathbf{Z})\) statistic.

However, although intuitively these studies seem to expand or shrink the rejection region through the term \(p\), directly comparing \(G^2(X,Y|\mathbf{Z}) - p\) with \(\chi^{2}_{\alpha,f}\) is not reasonable. This is because the \(G^2\)-statistic has a specific asymptotic distribution theory, and subtracting a value could distort its original statistical properties. The new statistic may no longer asymptotically follow a chi-squared distribution, thus invalidating hypothesis tests based on the chi-squared distribution. Even if \(G^2(X,Y|\mathbf{Z}) - p\) can be transformed to follow a chi-squared distribution, the related statistical properties and critical values, such as the degrees of freedom \(f\), need to be derived anew. Therefore, unreliable priors, when introduced into conditional independence testing, will also break the theoretical guarantees.

\section{Defining the Functional Boundaries of LLMs in Collaboration with CDAs}
\label{sec5}

\subsection{Functional Boundaries Illustration}

Despite the limitations of LLMs in reliably identifying causal relationships, this does not render them entirely useless for causal discovery tasks. Through extensive pretraining on vast textual data, LLMs accumulate substantial world knowledge, encompassing various domain-specific concepts, facts, and relationships. This extensive knowledge can provide intuitive guidance for causal discovery by facilitating more efficient exploration of causal structures. However, their role must be strictly confined to \textbf{a non-decisional auxiliary capacity}. The core principle is:
\begin{position}
\label{pos_5}
  In the collaboration between CDAs and LLMs, LLMs should not participate in determining the existence or directionality of causal relationships but can influence the search procedure for causal structures, thereby expediting the optimization process.
\end{position}
Specifically: (1) Prohibited Actions: LLM outputs must not serve as the final criterion for causal structures, such as directly deciding whether an edge exists between variables, the direction of edges, or acting as the core weight in scoring functions to influence decisions. (2) Permitted Actions: LLMs can be used for intermediate steps like initializing the search space, guiding mutation directions in evolutionary algorithms, or assisting with cycle detection. However, the final causal structure must rely entirely on reliable techniques, e.g., CDA scoring functions or constraint tests.

\subsection{Case Study: Use LLM to Guide Heuristic Search in CDAs}

To substantiate Position \ref{pos_5}, we present a concrete case study in this section: employing LLMs to guide heuristic search. Heuristic search is a commonly used strategy in causal discovery \cite{larranaga2013review}, including methods such as hill climbing and evolutionary algorithms \cite{chen2022genetic}. These techniques leverage heuristic information to enhance search efficiency. Traditional heuristic search methods often rely on random exploration during iterative optimization, which is inefficient and prone to local optima. In this section, we replace random search with LLM-driven targeted search while retaining classical scoring functions (e.g., BIC and BDeu scores) as optimization objectives. By integrating LLMs' knowledge and reasoning capabilities into the heuristic search process, search algorithms can intelligently select search directions, thereby accelerating convergence to the optimal solution. We incorporate LLMs into heuristic search from three perspectives:

(1) \textbf{LLM-Based Initial Population Initialization}: Before initiating heuristic search, LLMs analyze the causal structure space preliminarily. Given a causal learning dataset, variable information and their background knowledge are input into the LLM, which assesses their potential relationships. 
Variable pairs deemed highly unlikely to have causal relationships by the LLM are pruned from the search space at the initialization stage, significantly reducing the complexity. This well-informed initialization allows heuristic search to commence from a relatively accurate and sparse DAG, enhancing initial search efficiency.

(2) \textbf{LLM-Guided Evolutionary Optimization}: LLM is used to guide population evolution, as crossover and mutation operations. During mutation, given the current causal structure, LLMs receive detailed structural information along with mutation objectives (e.g., increasing diversity or refining local structures) and generate plausible mutation proposals. Unlike traditional random mutations, LLMs suggest where to add or remove edges based on their domain knowledge. Similarly, during crossover, LLMs evaluate two parent causal structures and recommend reasonable crossover points and strategies, ensuring offspring inherit desirable traits while exploring new structural spaces.

(3) \textbf{Cycle Detection and Resolution}: During heuristic search, generated causal structures may contain cycles, violating the acyclic constraint of DAGs. We employ LLMs for cycle detection and resolution. Once a new causal structure is formed and cycles are detected, its information is input into the LLM, where node and edge relationships are analyzed. LLMs suggest edges to remove or adjust based on their understanding of variable dependencies, ensuring the resulting causal structure remains a valid DAG.

We present the experimental results on bnlearn datasets of different scales in \textbf{Appendix} \ref{app_exsec3_5}, demonstrating the performance advantages of LLM-guided causal discovery. We also introduce other potential extensions of LLM-assisted evolutionary optimization for CDA in \textbf{Appendix} \ref{app_exsec3_5}, including the dynamic adjustment of search strategies, parameters, and uses accumulated search information to predict regions with a higher chance of having optimal solutions.

\section{Alternative Views Based on ``Manipulability of Experimental Results''}
\label{sec6}

Actually, most of the studies listed in \textbf{Appendix} \ref{app_summary} support the application of LLM in causal discovery. These studies typically prove their points from the perspective of empirical research on small-scale networks (with the number of nodes less than 100), and indicate that LLMs can provide relatively precise causal relationship identifications \cite{jincan,kiciman2023causal,jin2023cladder}. Some studies also support the collaboration of LLM and traditional CDA, and claim that even when LLMs yield incorrect results, score-based methods driven by soft constraints can still guide the CDA toward producing correct outputs \cite{ban2023causal,ban2023query,ban2025llm}. These findings appear to contradict the claim made in our position that ``LLMs cannot discover causality." Next, we first demonstrate why existing studies can still obtain favorable experimental results even though LLMs cannot identify causal relationships. Subsequently, we conduct a comprehensive experimental comparison among existing LLM-based causal discovery methods, traditional CDAs, and our proposed LLM-based heuristic search, highlighting the differences between these approaches.

\textbf{Inflated Performance Outcomes via Prompt Engineering}: As analyzed in Section \ref{sec3}, LLMs’ causal identification accuracy heavily relies on prompt-provided textual information, especially explicit background knowledge about variable relationships. Carefully designed prompts can artificially inflate performance—for instance, when prompts explicitly state known causal links (e.g., "Gene X causes Disease Y" from medical literature), LLMs merely retrieve preexisting knowledge rather than discover novel causality. This setup bypasses genuine causal inference, as the goal of causal discovery is to uncover unknown relationships, not replicate known ones. While LLMs can parrot causal statements under such conditions, their conclusions lack reliability, being neither consistently correct nor theoretically grounded. In global causal network discovery, incorporating LLM-generated priors into scoring functions (Section \ref{sec4}) exhibits a similar paradox: flawed methodologies yield empirically improved results. This stems from two factors: (1) Soft-constrained scoring functions, though theoretically unsound, may coincidentally align numerically when prior and data terms are scaled similarly; (2) In experiments (e.g., on `bnlearn` datasets), researchers often leverage ground-truth knowledge in prompts to manually engineer favorable outcomes, particularly on small-scale networks where prompt refinement is feasible.

To validate our argument, we experimented with soft-constrained scoring functions using GPT-4 on 14 small datasets from `bnlearn' in \textbf{Appendix} \ref{app_exsec3_6}. We tested two prompt types: (1) High-quality prompts, manually curated and refined to ensure accurate causal statements with minimal redundancy; (2) General prompts (Low-quality), sourced from online repositories like Wikipedia, containing redundant and potentially incorrect causal information (see Figure \ref{app_ex_6_1}). Results show that despite varying levels of incorrect prior knowledge, soft-constrained scoring functions consistently struggle to filter out errors. Under low-quality prompts, most erroneous priors from LLMs persist. While improving prompt quality reduces errors, this improvement stems from manual refinement rather than the effectiveness of soft constraints, highlighting their limitations.

\section{Conclusion: A Call to the Community}

This paper challenges the prevailing enthusiasm for using LLMs in causal discovery, revealing fundamental limitations in their ability to identify causal relationships and the risks of integrating them into CDAs. However, our critique does not dismiss LLMs entirely. We propose a cautious, theory-preserving role for LLMs as heuristic guides in causal structure search. Based on the discussion, we present the call to the community:
\begin{enumerate}[leftmargin=2em]
\item \textbf{Exercise Caution with LLM-Based Causal Discovery Methods}: Practitioners must recognize that current LLM-based algorithms for causal relationship identification lack theoretical guarantees and may produce unreliable results.
\item \textbf{Re-Evaluate Experimental Designs}: Many existing studies demonstrating ``successful'' LLM-driven causal discovery may inadvertently benefit from prompt engineering or information leakage (e.g., ground-truth knowledge in prompts). The community should adopt transparent, unbiased evaluation frameworks that exclude such artifacts, focusing on scenarios where causal relationships are genuinely unknown.  
\item \textbf{Prioritize Theoretical Rigor in Method Development}: Integrating LLMs as priors or decision components in CDAs should not circumvent the field’s foundational requirement for theoretical soundness. If LLMs are to be used, their role should be strictly confined to non-decisional tasks (e.g., search acceleration) to preserve CDA guarantees. Researchers should resist the temptation to justify LLM use with ad-hoc rationales that overlook their fundamental limitations.  
\item \textbf{Invest in LLM Architecture and Training for Causal Reasoning}: The community should explore how causal discovery can benefit from LLMs from the perspectives of LLM architecture and training process. It is not advisable to force general-purpose LLMs to identify causal relationships through prompt engineering without making any modifications to LLMs as in existing methods.
\end{enumerate}



\bibliography{example_paper}
\bibliographystyle{unsrtnat}

\newpage
\appendix


\section{Summary of Existing Studies for LLM-based Causal Discovery}
\label{app_summary}

Causal discovery is a fundamental task in causal inference \cite{glymour2019review}, aiming to uncover underlying causal relationships from observational data. Traditional causal discovery methods, including constraint-based and score-based approaches \cite{vowels2022d}, are grounded in statistical analysis and mathematical reasoning, providing rigorous and interpretable frameworks for inferring causal structures. While these methods have been widely applied across various domains, recent advancements in LLMs have introduced new opportunities for causal discovery. Unlike purely data-driven approaches, LLM-based methods can leverage rich linguistic knowledge, common sense reasoning, and domain-specific insights to complement traditional techniques. To provide a comprehensive overview, we summarize existing methods in Table \ref{tab_summary}. Based on these studies, we introduce current approaches along three key dimensions: the form of causal relationships considered, the design of prompts, and the integration strategies between LLMs and traditional algorithms.

\textbf{Pairwise Causality and Global Causal Graph}: Existing LLM-based causal discovery methods can be broadly categorized into two main groups: those focused on pairwise causal relationships and those aimed at learning the full causal graph. The key distinction between these two categories lies in their scope and technical approach. The first category of LLM-based methods concentrates on inferring the causal direction between a pair of variables (A$\leftarrow$B or A$\rightarrow$B) or determining the existence of a causal relationship \cite{kiciman2023causal,zevceviccausal}. These approaches leverage the language understanding and reasoning capabilities of LLMs to make judgments about the causal connections between individual variable pairs. The key idea is to prompt the LLM with the variable names, potentially along with some contextual information, and have the model output its assessment of the causal relationship. Some studies have shown that LLMs like GPT-4 can outperform traditional CDAs in some tasks \cite{jincan}, but other studies also recognized the limitations that LLMs may simply repeat patterns embedded in their training data, rather than engaging in genuine causal reasoning \cite{chiunveiling}. In contrast, the second category of LLM-based methods aims to uncover the complete causal graph, involving all the variables in the system \cite{long2023causal,ban2023causal}. These approaches go beyond pairwise relationships and seek to model the overall causal structure. Typically, these methods combine the language understanding of LLMs with data-driven CDAs, where the LLMs are used to generate causal insights or knowledge that can then be integrated with numerical reasoning techniques. The LLMs' ability to reason about the contextual information and leverage domain knowledge plays a crucial role in these full graph discovery tasks, which are generally more complex than the pairwise setting.

\textbf{Prompt Design for Identifying Causality:} The design of the prompt plays a critical role in guiding LLMs' causal reasoning capabilities in both pairwise and full causal graph discovery tasks. Researchers have explored incorporating various types of information into the prompts to provide the necessary context and guidance for the LLMs: (1) Variable information \cite{constantinou2024using}: By explicitly specifying the variables involved, the LLMs can better understand the entities and their potential causal relationships. (2) Background knowledge \cite{zhang2023understanding}: This background knowledge can range from simple problem descriptions to more detailed information from scientific literature \cite{zhang2024causal}. The rationale behind this is to equip the LLMs with relevant domain expertise and contextual understanding, which can provide them with information and context they may not have encountered during the pre-training stage. (3) Observational data \cite{cai2023knowledge}: By exposing the LLMs to the actual data, either in the form of raw data or summarized statistics, the models can leverage the empirical information to inform their causal reasoning, potentially leading to more grounded and reliable conclusions. (4) Results of CDA \cite{abdulaal2023causal}: In the case of full causal graph discovery, researchers have also explored including intermediate causal discovery results in the prompts. This approach aims to guide the LLMs' reasoning by providing them with partial causal insights, which the models can then use to refine and expand the overall causal structure. The underlying idea is to leverage the LLMs' language understanding capabilities to integrate these intermediate results and generate a more comprehensive causal graph.

\textbf{Collaboration Paradigm between LLMs and CDAs:} Recent studies have explored hybrid approaches that integrate LLMs with traditional CDAs. These hybrid techniques seek to leverage the complementary strengths of both LLMs and established numerical techniques. Based on the different ways these two components are combined, these hybrid methods can be broadly categorized into four main types: 
(1) Initialization or Prior: LLMs are used to generate initial causal insights or hypotheses, which are then used to initialize or provide a prior for the traditional CDAs. This allows the algorithms to start from a more informed state \cite{ban2023query,darvariu2024large,naik2024applying}. 
(2) Post-processing: The traditional CDAs are first applied, and the results are then further refined or expanded by the LLMs. The LLMs can leverage their language understanding and reasoning capabilities to build upon the outputs of the traditional methods \cite{long2023causal,castelnovo2023marrying,khatibi2024alcm}. 
(3) Iterative Collaboration: The LLMs and traditional algorithms are used in an iterative, back-and-forth manner. The models can exchange information, with the LLMs providing contextual knowledge or insights that guide the traditional algorithms, and the algorithms offering numerical reasoning to validate or refine the LLMs' outputs \cite{abdulaal2023causal,ban2023causal,takayama2024integrating}. 
(4) Modular Integration: LLMs can be used to implement specific modules or components within the traditional CDAs, such as scoring functions or conditional independence tests. This allows the LLMs to contribute their strengths to the overall causal inference process \cite{zhang2024causal,cohrs2024large}.

\begin{table}[H]
\caption{Overview of Causal Discovery Research Based on LLM (Part 1)}
\label{tab_summary}
\centering
\resizebox{1\textwidth}{!}{ 
\begin{tabular}{p{3cm}p{1.8cm}p{3.4cm}p{5cm}p{1.6cm}}
\toprule
Research & Form of Identified Causality & Prompt Content & How to Combine with Causal Discovery Algorithm (CDA) & Maximum DAG in Experiment \\
\midrule
\cite{zhiheng2022can} & Pairwise & Variable Information, Background Knowledge & N/A & 2 \\
\cite{long2022can} & Pairwise & Variable Information, Causal Statement among Variables & N/A & 4 \\
\cite{zhang2023understanding} & Pairwise & Variable Information, Background Knowledge & N/A & N/A \\
\cite{willig2023probing} & Pairwise & Variable Information, Background Knowledge & N/A & 6 \\
\cite{kiciman2023causal} & Global DAG & Variable Information & N/A & 12 \\
\cite{long2023causal} & Global DAG & Variable Information & Use CDA to find Markov equivalence classes and use LLM to obtain the DAG & 37 \\
\cite{ban2023query} & Global DAG & Variable Information & Take the LLM's results as prior and incorporate into a score-based CDA & 48 \\
\cite{chen2023mitigating} & Global DAG & Variable Information & Take the LLM's results as prior and incorporate into a score-based CDA & 37 \\
\cite{zevceviccausal} & Pairwise & Variable Information, Background Knowledge & N/A & 6 \\
\cite{abdulaal2023causal} & Global DAG & Variable Information, Background Knowledge, CDA Results & Initialize with LLM, then obtain a DAG using CDA, and finally refine the DAG using LLM & 15 \\
\cite{gao2023chatgpt} & Pairwise & Variable Information, Causal Statement among Variables & N/A & N/A \\
\cite{castelnovo2023marrying} & Global DAG & Variable Information & Use CDA to obtain an initial DAG and use LLM to refine the DAG & N/A \\
\cite{ban2023causal} & Global DAG & Variable Information, Background Knowledge & Iteratively use the CDA and LLM & 84 \\
\cite{jin2023cladder} & Pairwise & Variable Information, Causal Statement among Variables & N/A & 4 \\
\cite{jincan} & Pairwise & Variable Correlation & N/A & 6 \\
\cite{cai2023knowledge} & Pairwise & Variable Information, Observation Data & N/A & 12 \\
\cite{jiralerspong2024efficient} & Global DAG & Variable Information, Variable Correlation & N/A & 221 \\
\cite{takayama2024integrating} & Global DAG & Variable Information, CDA Results & Iteratively use the CDA and LLM & 11 \\
\cite{tong2024automating} & Pairwise & Variable Information, Background Knowledge (Scientific Publication) & N/A & N/A \\
\cite{zhang2024causal} & Global DAG & Variable Information, Background Knowledge (Scientific Publication) & Use a constraint-based CDA and leverages LLMs to perform conditional independence queries & 11 \\
\cite{liu2024llms} & Pairwise & Variable Information, Observation Data, Background Knowledge & N/A & 25 \\
\cite{chiunveiling} & Pairwise & Variable Information, Background Knowledge & N/A & 2 \\
\cite{khatibi2024alcm} & Global DAG & Initial DAG, Variable Information, Background Knowledge & Use CDA to obtain an initial DAG and use LLM to refine the DAG  & 221 \\
\bottomrule
\end{tabular}
}
\end{table}

\begin{table}[t]
\caption{Overview of Causal Discovery Research Based on LLM (Part 2)}
\centering
\resizebox{\textwidth}{!}{ 
\begin{tabular}{p{3cm}p{1.8cm}p{3.4cm}p{5cm}p{1.6cm}}
\toprule
Research & Form of Identified Causality & Prompt Content & How to Combine with Causal Discovery Algorithm (CDA) & Maximum DAG in Experiment \\
\midrule
\cite{darvariu2024large} & Global DAG & Initial DAG, Variable Information & Take the LLM's results as prior and incorporate into a score-based CDA  & 27 \\
\cite{naik2024applying} & Global DAG & Variable Information, Background Knowledge & Take the LLM's results as prior and then use CDA  & 18 \\
\cite{cohrs2024large} & Global DAG & Variable Information, Background Knowledge & Use a constraint-based CDA and leverages LLMs to perform conditional independence queries  & 11 \\
\cite{le2024multi} & Global DAG & Variable Information, Background Knowledge, Observation Data & N/A  & N/A \\
\cite{constantinou2024using} & Global DAG & Variable Information & N/A & 56 \\
\cite{chen2024causal} & Global DAG & Variable Information & N/A & 9 \\
\cite{li2024realtcd} & Global DAG & Variable Information, Background Knowledge & Use LLM to obtain an initial DAG and use a score-based CDA to refine the DAG & 38 \\
\cite{zhang2024causalchat} & Pairwise & Variable Information, Background Knowledge, CDA Results & N/A & 9 \\
\cite{kampani2024llm} & Global DAG & Variable Information, CDA Results & Take the LLM's results as prior and then use a CDA & 70 \\
\cite{ban2025llm} & Global DAG & Variable Information, Background Knowledge & Take the LLM's results as prior and then use a CDA & 37 \\
\cite{meier2025structured} & Global DAG & Variable Information, Background Knowledge &  N/A & 100 \\
\cite{wang2025large} & Global DAG & Variable Information &  Take the LLM's results as prior and incorporate into a score-based CDA & 413 \\
\cite{vashishtha2025causal} & Global DAG & Variable Information, Background Knowledge &  Take the LLM's results as prior and then use CDA & 22 \\
\cite{cai2025dr} & Pairwise & Variable Information, Background Knowledge & N/A  & N/A \\
\cite{yu2025causaleval} & Global DAG & Variable Information, Background Knowledge, Structured Data & Take the LLM's results as prior and then use CDA  & 5 \\
\cite{li2025can} & Global DAG & Variable Information, Background Knowledge & Use LLM to guide the selection of intervention targets in active CDA.  & 37 \\
\cite{ban2025integrating} & Global DAG & Variable Information, Background Knowledge & Take the LLM's results as prior and incorporate into a score-based CDA  & 48 \\
\cite{shukla2025kulfi} & Pairwise & Variable Information, Background Knowledge & N/A  & N/A \\
\cite{zanna2025fairness} & Global DAG & Variable Information, Background Knowledge & N/A  & 221 \\
\cite{cohrs2025large} & Global DAG & Variable Information, Background Knowledge & Use a constraint-based CDA and leverages LLMs to perform conditional independence queries  & 11 \\
\cite{susanti2025can} & Pairwise and Global DAG & Variable Information, Observation Data, Background Knowledge & N/A & 8 \\
\cite{newsham2025large} & Global DAG & Variable Information, Background Knowledge (Scientific Publication) & N/A & 100 \\
\cite{li2025beyond} & Global DAG & Textual Data & N/A & 40 \\
\cite{sheikh2025comparing} & Global DAG & Variable Information, Background Knowledge & N/A & N/A \\
\bottomrule
\end{tabular}
}
\end{table}

\section{Analysis and Empirical Studies of Position \ref{pos_1}}
\label{app_exsec3_1}

This appendix provides a systematic analysis and empirical studies of the causality identification capabilities of LLMs, further validating and quantifying the key influencing factors discussed in Section \ref{sec3}. Specifically, we designed multiple experiments to examine the impact of word order, redundant information, and the distance between causal entities on the causal reasoning ability of LLMs.

\subsection{Analysis of Factor (1) in Position \ref{pos_1}}

First, it is essential to ensure that the causal relationships in the context are clear and familiar. Starting from the autoregressive mechanism, the representation form of causal relationships and the word order in the prompt should be similar to the causal patterns that frequently appear in the training corpus. The example of ``rain" mentioned in Section \ref{sec3} can intuitively reflect the motivation for this requirement. To verify the impact of word order in the text on the ability of LLMs to identify causal relationships between entities, we tested the performance of several mainstream LLMs, including GPT-3.5, GPT-4o, Claude 3.5, Llama 3.1, Gemini 1.5, and Falcon-40B, using different sentence patterns and word orders. Specifically, we extensively collected test examples from large - scale general - purpose corpora (such as Wikipedia, news corpora, etc.) and corpora in specific fields (such as medical and scientific literature). The expressions of causal relationships are divided into two categories. One category consists of words that explicitly indicate causal relationships, such as: 
\begin{tcolorbox}
Because, Since, Therefore, Consequently, Thus, Hence, As a result, Due to, Owing to, Resulting in, For this reason, So, Caused by, Cause - Effect.
\end{tcolorbox}
The other category consists of words that implicitly suggest causal relationships, such as: 
\begin{tcolorbox}
Trigger, Lead to, Bring about, Give rise to, Imply, Influence, Promote, Encourage, Contribute to, Pave the way for, Stem from, Derived from, Necessitate.
\end{tcolorbox}

\begin{figure}[t]
  \centering
  \includegraphics[width=12cm]{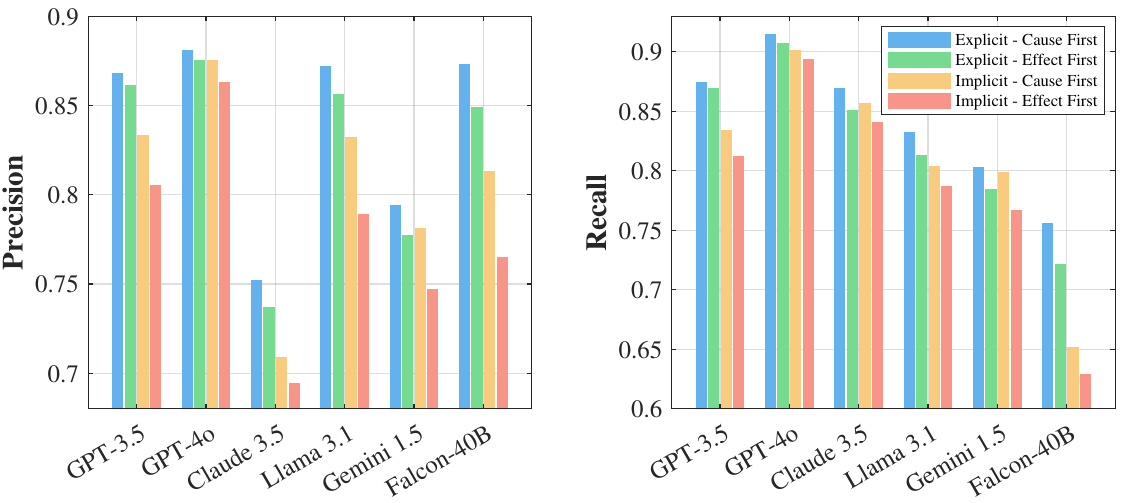}
  \caption{The precision and recall of different LLMs under impact of causal relationship expression.}\label{app_ex_1}
\end{figure}

Based on these causal words, we constructed a series of text sets containing causal relationships. For each text, we rewrote it to generate two versions: one with the cause preceding the effect and the other with the effect preceding the cause. Meanwhile, we added redundant information and intermediate events to increase the complexity of the task. We also detected the detection accuracy of LLMs for true and false causal relationships, and Figure \ref{app_ex_1} shows the precision and recall of each LLM under different conditions. Through testing with different sentence patterns and word orders, we observed several notable trends. Firstly, LLM generally perform better on texts with explicit causal relationships compared to texts with implicit causal relationships, which indicates that LLMs can more effectively identify and understand causal relationships when dealing with clear causal words. Secondly, when we adjusted the order of entities, that is, placing the effect before the cause, the performance of some LLMs significantly decreased, especially in texts with implicit causal relationships. This result shows that LLMs are sensitive to the order of word and may rely on common causal patterns in training for reasoning. Therefore, the structure of the text plays a crucial role in the model's understanding process. Finally, by comprehensively analyzing the performance of various models, we found that models with higher recall often have lower precision. This phenomenon indicates that these LLMs have certain inclinations. In some cases, such as GPT-4o, it is more inclined to judge "there is a causal relationship", while in other models, such as Falcon-40B, it is more inclined to judge "there is no causal relationship". This inclination, derived from the training data, may reflect the inherent biases of the models in causal reasoning, reminding us to be cautious when interpreting their discrimination results in practical applications.

\subsection{Analysis of Factor (2) in Position \ref{pos_1}}

In addition to the special requirements for word order, there should not be too many entities or redundant information in the text, which can be understood from the perspective of information entropy. In causal learning, we are concerned with the mutual information \(I(\mathbf{x}_c;\mathbf{x}_e)\) between the cause and the effect, where \(\mathbf{x}_c\) and \(\mathbf{x}_e\) represent the sets of words related to the cause and the effect, respectively. Consider the conditional mutual information:
\begin{equation}\label{eq_3}
  I(\mathbf{x}_c;\mathbf{x}_e|\mathbf{x}_z)=H(\mathbf{x}_c|\mathbf{x}_z)+H(\mathbf{x}_e|\mathbf{x}_z)-H(\mathbf{x}_c,\mathbf{x}_e|\mathbf{x}_z)
\end{equation}
where \(\mathbf{x}_z\) represents other information in the text. Excessive entities and redundant information make \(P(\mathbf{x}_c,\mathbf{x}_e|\mathbf{x}_z)\) more dispersed, resulting in an increase in the joint entropy \(H(\mathbf{x}_c,\mathbf{x}_e|\mathbf{x}_z)\), forcing entities that are far apart in the text to tend to have no causal relationship. At the same time, redundant entities make the conditional entropies \(H(\mathbf{x}_c|\mathbf{x}_z)\), \(H(\mathbf{x}_e|\mathbf{x}_z)\) and \(H(\mathbf{x}_c,\mathbf{x}_e|\mathbf{x}_z)\) more complex, making it difficult for the model to accurately grasp the mutual information between the cause and the effect given these interfering information, thus increasing the difficulty of understanding causal relationships from the text.

To investigate the impact of redundant entities on the ability of LLMs to recognize causal relationships in text, we continued to test several widely popular LLMs, including GPT-3.5, GPT-4o, Claude 3.5, Llama 3.1, Gemini 1.5, and Falcon-40B. First, we identified a series of simple causal relationship examples as the basic texts, with the form such as ``Because (Entity 1), so (Entity 2)". Each basic text clearly expressed a causal relationship and was easy to understand. Different types of basic texts covered multiple fields such as life, nature, and society to ensure the comprehensiveness of the experiment. Then, we added redundant information to each basic text. The main positions for adding redundant information included the following four types:
\begin{itemize}
  \item Redundancy before the first entity: Adding redundant text or entities before the first entity of the basic text.
  \item Redundancy between two entities: Inserting redundant content between the first and the second entities of the basic text.
  \item Redundancy after the second entity: Adding redundant parts after the second entity of the basic text.
  \item Redundancy at both ends: Adding redundant information before the first entity and after the second entity of the basic text simultaneously, but not between the two entities.
\end{itemize}

\begin{figure}[t]
  \centering
  \includegraphics[width=8cm]{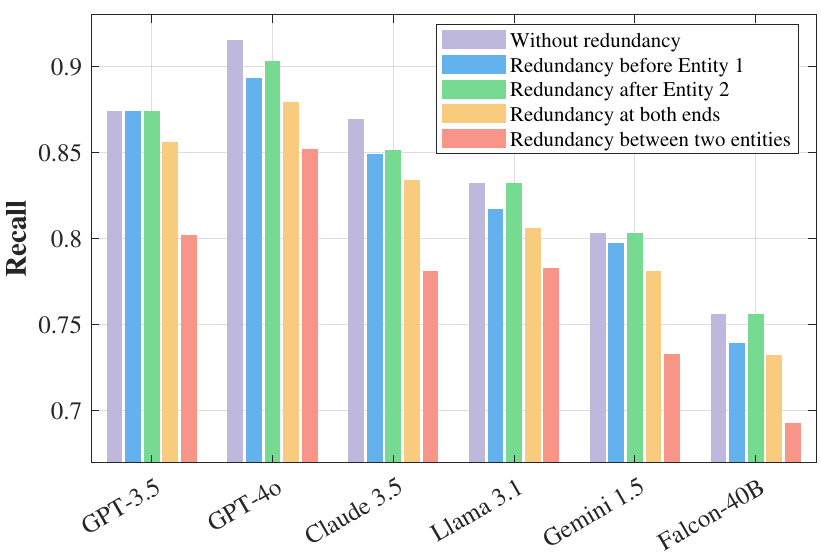}
  \includegraphics[width=8cm]{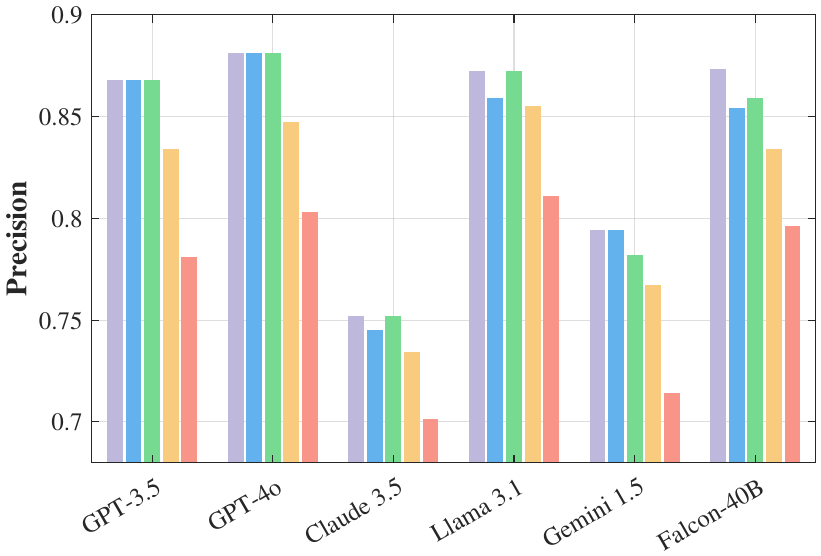}
  \caption{The precision and recall of different LLMs under impact of redundant expression.}\label{app_ex_2}
\end{figure}

According to the above designed method, a large number of test texts with different redundant positions and different redundancy measures were generated. In the comparison, similar numbers of words and the same number of entities were added at different redundant positions. For each selected LLM, the texts in the test set were successively input into the model, and the recognition results of the causal relationships in the text by the model were obtained. When calling the model, the input format and questioning method were kept consistent. The causal relationship recognition results output by the model were compared with the pre-annotated correct causal relationships. For each test text, it was recorded whether the model correctly recognized the causal relationship and whether it wrongly recognized a non-existent causal relationship. The recall and precision rates of each model under different redundant positions were calculated separately, and the results are shown in Figure \ref{app_ex_2}.

Overall, when redundant information was added separately in the front and back of the causal text, the impact on the precision and recall rates of each model was not significant. However, when comparing the front and back positions, adding redundant information before the first entity had a relatively greater impact on the model performance than adding it after the second entity. When redundant information was added simultaneously at the front and back of the causal text, the precision and recall rates of each model decreased to a certain extent. This, to some extent, reflects that LLMs tend to pay more attention to the information at the beginning and end of the text when processing it. When the key causal information is in the middle of the text, it is easily ignored by the model due to the interference of redundant information at the front and back, thus affecting its accurate recognition of causal relationships. Adding redundant information between the two entities had a great impact on the ability of each model to recognize causal relationships. This is because the redundant information at this position widens the distance in the text space between the two causally related entities that were originally closely associated. LLMs usually process text based on the autoregressive mechanism of word order, and this mechanism is significantly affected when the text-based association between entities is weakened. These results fully demonstrate that the ability of LLMs to recognize causal relationships from text is affected by redundant information in the text. This, from the side, reflects that the process of LLMs recognizing causal relationships is not really about excavating the internal causal mechanisms between entities in the text, but rather more about summarizing the surface information of the text. When processing text, LLMs rely on the word order and information distribution presented in the text. When redundant information changes the text structure and information layout, the model's judgment of causal relationships will be interfered with, making it difficult to accurately grasp the true causal logic behind the text.

\subsection{Analysis of Factor (3) in Position \ref{pos_1}}

From Figure \ref{app_ex_2}, we discover that adding redundant information between the two entities had a great negative impact, which aligns with the Factor (3) in Position 3.1. Due to the technical characteristics of word embedding techniques and the attention mechanism, words that are adjacent in text positions usually have stronger semantic and syntactic associations. Therefore, the embedding vectors corresponding to tokens that are far apart and have no obvious semantic associations tend to be orthogonal (according to the properties of high-dimensional vectors), and correspondingly, the attention scores will be lower. On the other hand, from the perspective of information transfer, when calculating the degree of attention of one word to other words, the information carried by words that are far apart will have a smaller contribution weight to the current word after Softmax normalization. If the entities to be analyzed are far apart in the text, the LLM will have difficulty capturing the causal relationship between the two entities. 

\begin{figure}[t]
  \centering
  \includegraphics[width=6cm]{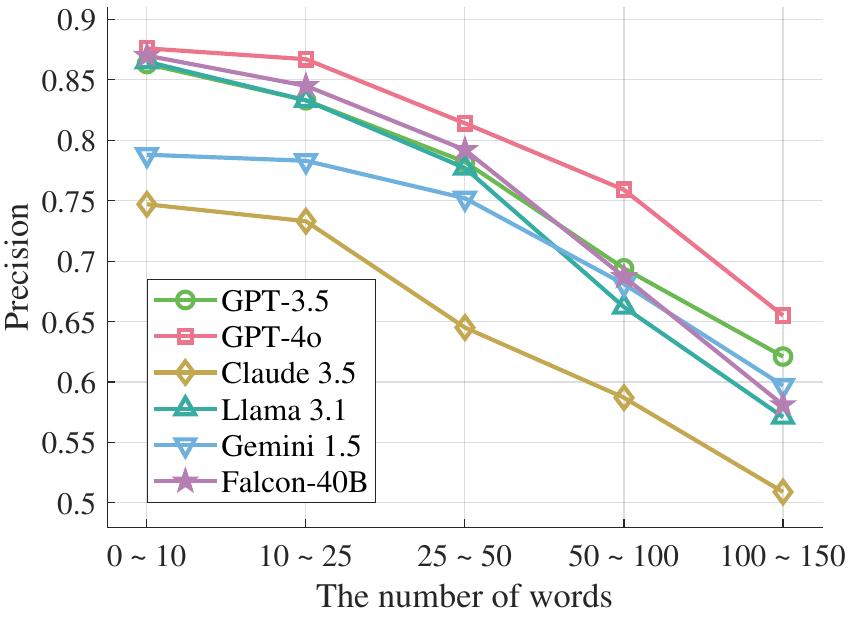}
  \includegraphics[width=6cm]{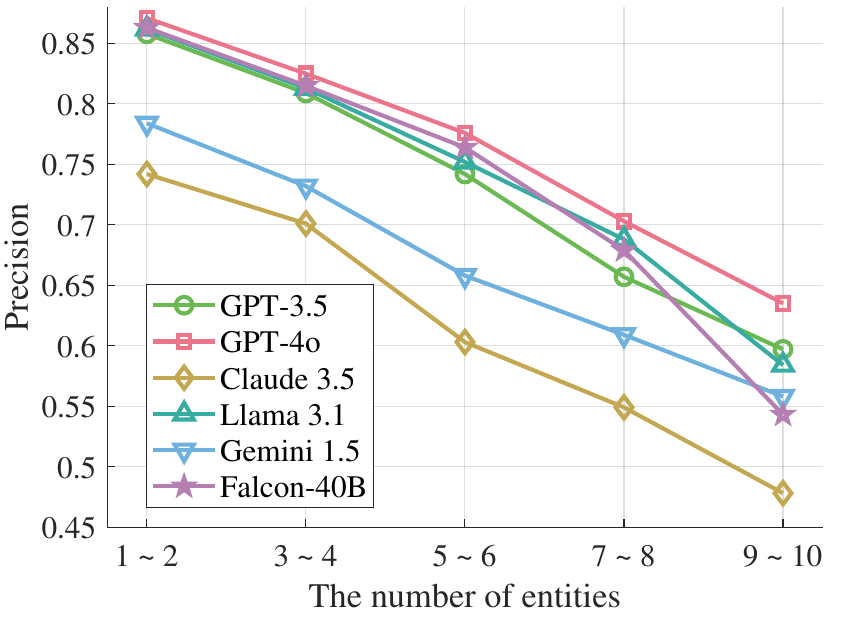}
  \includegraphics[width=6cm]{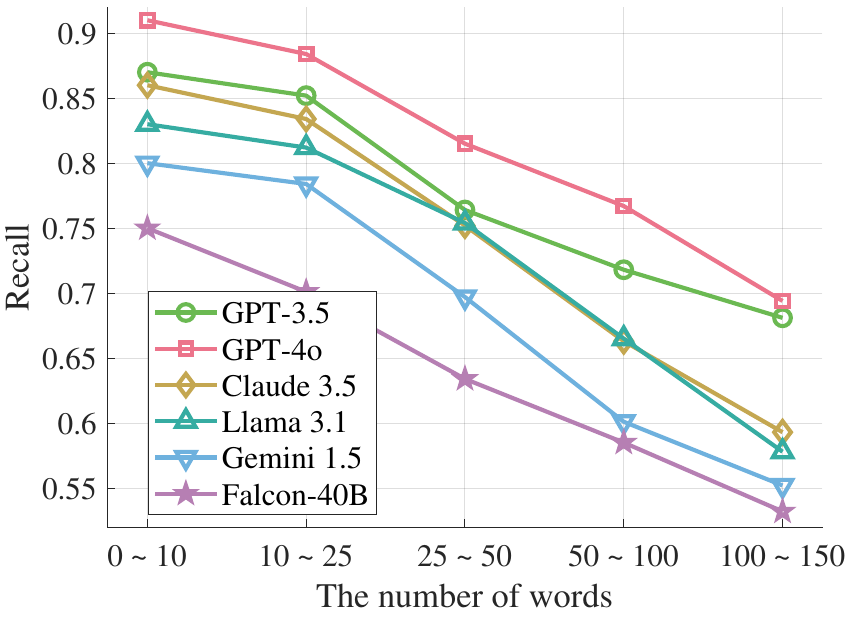}
  \includegraphics[width=6cm]{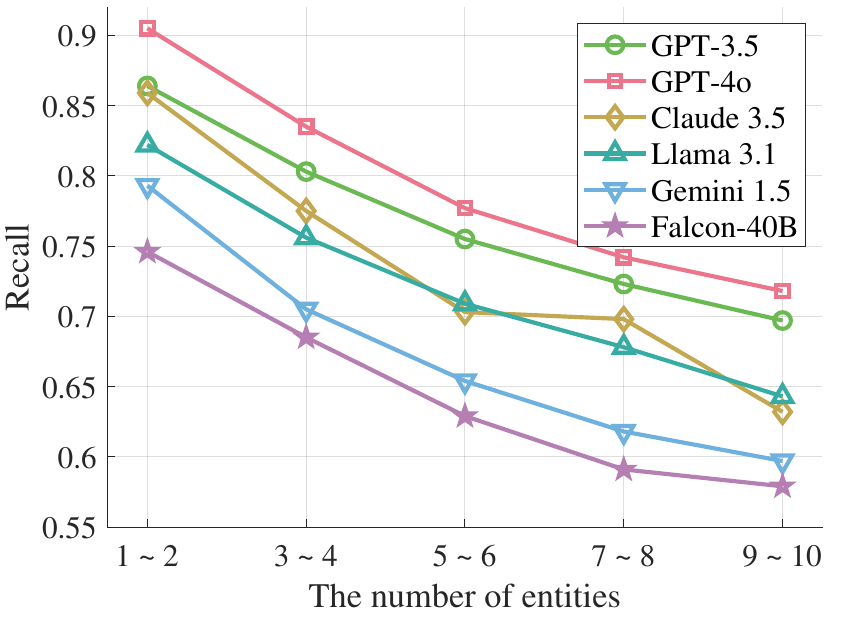}
  \caption{The precision and recall of different LLMs under impact of redundant expression.}\label{app_ex_3}
\end{figure}

To verify the impact of ``entity distance" under different measurement methods on the aforementioned LLMs' ability to identify causal relationships, we adopt two ways to measure the distance between two entities, based on the number of words and the number of entities respectively:
\begin{itemize}
  \item Based on the number of words: For each position where redundancy is added, redundant texts with word counts in the ranges of 0 - 10, 10 - 25, 25 - 50, 50 - 100, and 100 - 150 are added respectively, thereby generating different versions of test texts.
  \item Based on the number of entities: For each position where redundancy is added, redundant entities with the number of entities in the ranges of 1 - 2, 3 - 4, 5 - 6, 7 - 8, and 9 - 10 are added.
\end{itemize}

We select a variety of common causal relationship scenarios, such as weather and traffic, disease and symptoms, behavior and consequences, etc. In each scenario, two entities of the core causal relationship are determined. For example, in the weather-and-traffic scenario, ``sudden heavy snowfall" and ``traffic congestion" are taken as the causal entities. Subsequently, different numbers of words and other entities are added between the two core entities to control the lexical distance. The added content is used to describe intermediate events, background information, irrelevant details, etc., constructing text samples with different lexical distances. For each LLM, the texts in the test set are successively input into the model, and the model's recognition results of the causal relationships in the text are obtained. The causal - relationship recognition results output by the model are compared with the pre - annotated correct causal relationships, and the recall and precision rates of each model under different lexical distances are calculated. The experimental results are shown in Figure \ref{app_ex_3}.

From the results, it is not difficult to find that as the number of redundant words and redundant entities between the two entities increases, both the precision and recall of LLMs in identifying causal relationships gradually decline. Among them, GPT-4o has a continuous and significant advantage in recall compared with other LLMs, but its precision is similar to that of other LLMs. The decline in recall means that some distant causal relationships cannot be recognized. We find that as the number of redundant entities increases, the identification of causal relationships by LLMs gradually becomes random, and the results of multiple experiments are also unstable. On the other hand, the decline in precision means that more and more irrelevant entities are recognized as causal relationships, and we observe that these irrelevant entities are mostly distributed near the other entity. All these reflect that the entity distance has a significant impact on LLMs' identification of causal relationships.

\begin{figure}[t]
  \centering
  \includegraphics[width=12cm]{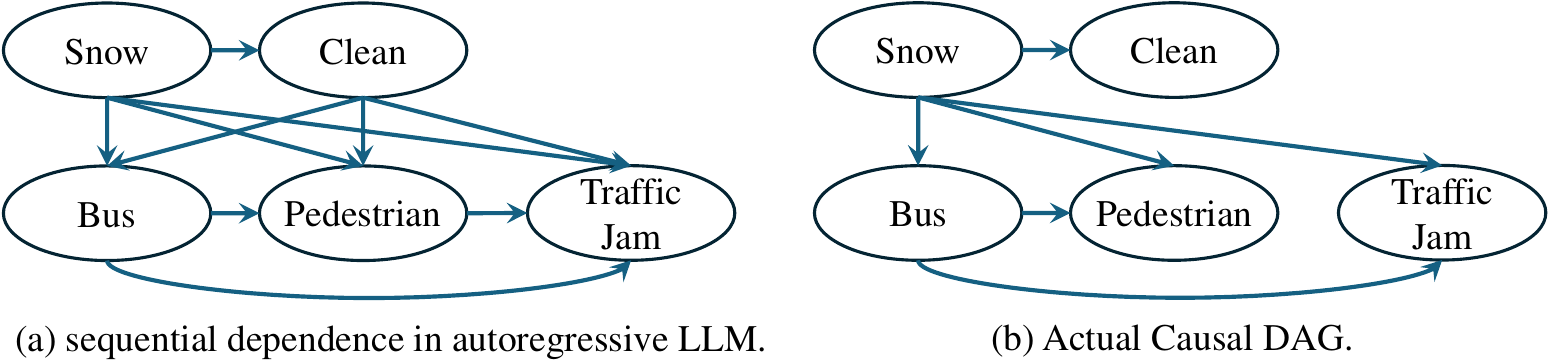}
  \caption{An example of Factors (2) and (3) in Position \ref{pos_1}.} \label{pic_sec3}
\end{figure}

In fact, for the requirements in (2) and (3), an example can be used to show the impact of text on the LLM. For example, consider the following text: 
\begin{tcolorbox}
Last night, there was a sudden heavy snowfall (Snow). The sanitation workers went out early in the morning to clear the snow on the main roads (Clean). The bus company temporarily adjusted some bus routes (Bus) to deal with possible road safety problems. Due to the slippery roads, pedestrians walked carefully (Pedestrian), and there was traffic congestion during the morning rush hour today (Traffic Jam).
\end{tcolorbox}
As shown in Figure \ref{pic_sec3}, because the autoregressive model reasons based on sequential dependence, the LLM sees the model in Figure \ref{pic_sec3}(a). But in fact, the node Snow is the direct cause of the node Traffic Jam (as shown in Figure \ref{pic_sec3}(b)). However, due to the many redundant events in the middle and the fact that Snow and Traffic Jam are far apart in the text, the LLM has difficulty capturing this complex actual causal relationship and is likely to regard intermediate events (such as Pedestrian) as the cause of Traffic Jam.

\section{Analysis and Empirical Studies of Position \ref{pos_2}}
\label{app_exsec3_2}

This appendix investigates the effects of numerical precision on LLMs' ability to understand correlation and causation in data and evaluates their performance using causal benchmarks. 

We begin by assessing the sensitivity of LLMs to numerical precision in the simplest task of recognizing correlations. Classic three-variable causal graphs are employed, with predefined causal mechanisms and distributions. Consider three variables \(X\), \(Y\), and \(Z\), connected in typical causal structures such as \(X \rightarrow Y \rightarrow Z\), \(X \leftarrow Y \rightarrow Z\), and \(X \rightarrow Y \leftarrow Z\). Data samples are simulated based on these causal graphs using Monte Carlo methods. Each variable's distribution is determined according to its causal relationships. For instance, if \(X\) is exogenous, it is sampled from a normal distribution \(N(\mu_1, \sigma_1^2)\). If \(Y\) is influenced by \(X\), it is generated as \(Y = f(X) + \epsilon_1\), where \(\epsilon_1 \sim N(0, \sigma_2^2)\). Similarly, \(Z\) is generated based on its causal relationships with \(X\) and \(Y\). A total of 100 samples are generated, each containing full observations of the three variables.

To explore the effect of numerical precision, we vary the level of precision in the observational data, retaining 1 to 8 significant digits through rounding. This simulates observational data of varying precision. These datasets are then formatted as prompts suitable for LLM input and tested on GPT-3.5, GPT-4o, Claude 3.5, Llama 3.1, Gemini 1.5, and Falcon-40B. The task requires models to identify correlations and conditional dependencies among variables, and their outputs are evaluated using accuracy metrics.

\begin{figure}[ht]
  \centering
  \includegraphics[width=7cm]{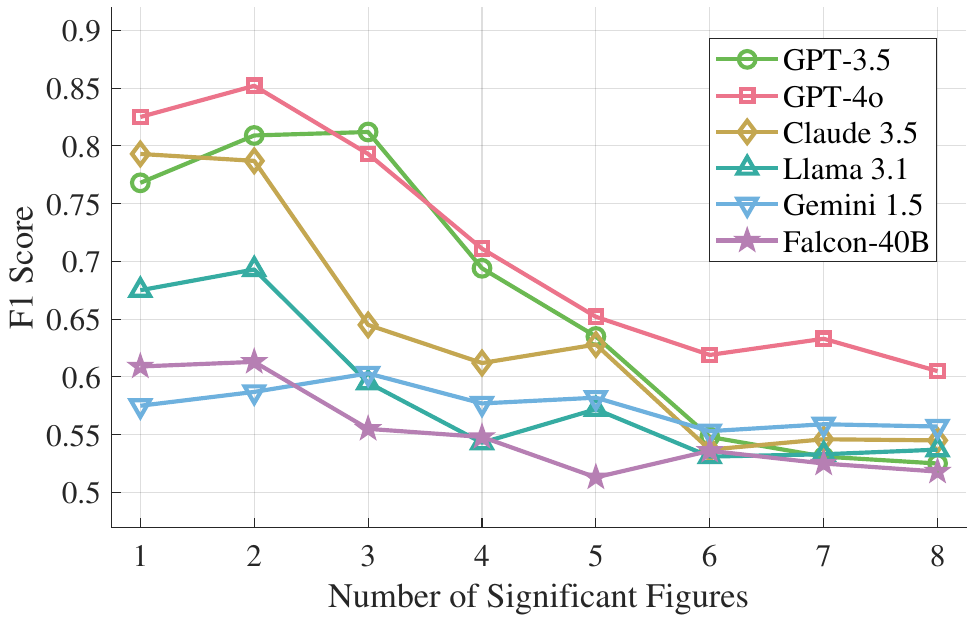}
  \caption{The F1 Score of different LLMs under different significant figures.} \label{ex_4}
\end{figure}

As shown in Figure~\ref{ex_4}, increasing numerical precision does not improve LLM performance as seen with traditional algorithms. Instead, LLMs exhibit a slight performance improvement at low precision (e.g., 1–2 significant digits) but degrade as precision increases. Ultimately, all LLMs achieve an F1 score near 0.5 at higher precision levels, indicating a performance equivalent to random guessing. This suggests that LLMs struggle to interpret high-precision numerical data. When numbers are encoded as tokens, the encoding process likely distorts critical numerical features, leading to information loss or misrepresentation. At lower precision levels, essential features are less affected by encoding, allowing models to leverage their language understanding and pattern recognition capabilities to identify causal relationships. However, as precision increases, the tokenized representation introduces excessive detail, hindering the extraction of key causal features.

To further evaluate LLMs' ability to infer causality from observational data, we utilize benchmark datasets from the \texttt{bnlearn} community~\cite{scutari2010learning}. These datasets span various domains, including healthcare, finance, and environmental sciences, and feature varying scales. For this experiment, we select 14 small-scale datasets with fewer than 100 nodes, as listed in Table~\ref{tab_ex5}. These datasets are well-suited for LLMs given their prompt length constraints. Observational samples from these datasets, which reflect variable relationships, are formatted as prompts and tested on GPT-3.5 and GPT-4, the two best-performing models from the previous experiment.

Performance is measured using two metrics: the F1 score and Structural Hamming Distance (SHD) \cite{norouzi2012hamming}. The F1 score captures the trade-off between precision and recall in identifying causal relationships. SHD measures the discrepancy between the inferred causal structure and the ground truth, accounting for added, reversed, and missing edges. Lower SHD values indicate more accurate graph structures. Experimental results, summarized in Table~\ref{tab_ex5}, show that GPT-4 significantly outperforms GPT-3.5. However, GPT-3.5 only succeeds in identifying causal relationships in the simplest datasets, such as the \textit{Earthquake} dataset, where observational data is beneficial. For higher-dimensional datasets, even GPT-3.5 struggles to extract meaningful causal information. Both models show declining performance with increasing dataset complexity. As dataset size grows, F1 scores decrease, indicating reduced precision and recall, while SHD values rise, reflecting greater structural differences from the ground truth. This trend underscores LLMs' limited capacity to extract causal relationships from observational data.

\begin{table*}[ht]
\label{tab_ex5}
    \centering
    \caption{The impact of the observational data.}
\resizebox{\textwidth}{!}{ 
    \begin{tabular}{@{}lccccccccc@{}}
        \toprule
        \textbf{Dataset} & \textbf{Node number} & \multicolumn{4}{c}{\textbf{GPT-3.5}} & \multicolumn{4}{c}{\textbf{GPT-4o}} \\ \cmidrule(lr){3-6} \cmidrule(lr){7-10}
                         &                       & \textbf{F1} & \textbf{F1$_D$} & \textbf{SHD} & \textbf{SHD$_D$} & \textbf{F1} & \textbf{F1$_D$} & \textbf{SHD} & \textbf{SHD$_D$} \\ \midrule
        Cancer          & 5                     & 0.5673      & 0.5449       & 7.4         & 8.6         & 0.6468      & 0.6712       & 6.8         & 6.4         \\
        Earthquake      & 5                     & 0.7202      & 0.7813       & 4.4         & 3.6         & 0.6485      & 0.7319       & 3.4         & 5.6         \\
        Survey          & 6                     & 0.5788      &   Fail           & 7.2         &   Fail           & 0.6024      & 0.5422       & 6.5         & 7.97        \\
        Asia            & 8                     & 0.5787      & 0.5148       & 18.4        & 24.2        & 0.5794      & 0.5988       & 18.2        & 21.8        \\
        Sachs           & 11                    & 0.4450       &   Fail           & 54.6        &  Fail            & 0.4836      & 0.4663       & 28.5        & 52.92       \\
        Child           & 20                    & 0.5018      &  Fail            & 62.1        &  Fail            & 0.4782      & 0.4766       & 83.6        & 84.1        \\
        Insurance       & 27                    & 0.4709      &  Fail            & 256         &  Fail            & 0.5505      & 0.5849       & 175         & 167.4       \\
        Water           & 32                    & 0.4855      &  Fail            & 68          &  Fail            & 0.4834      & 0.4119       & 68.6        & 159.8       \\
        Mildew          & 35                    & 0.4540       &   Fail           & 361.6       &   Fail           & 0.5452      & 0.5255       & 136.4       & 149.5       \\
        Alarm           & 37                    & 0.4740       &  Fail            & 347         &  Fail            & 0.4705      & 0.5126       & 406.4       & 232.1       \\
        Barley          & 48                    & 0.4862      &   Fail           & 437.2       &  Fail            & 0.4979      & 0.4765       & 312.4       & 145.41      \\
        Hailfinder      & 56                    & 0.3985      &  Fail            & 1178.8      &   Fail           & 0.3620       & 0.2959       & 1420.6      & 1865.2      \\
        Hepar II        & 70                    & 0.4797      &  Fail            & 1299        &  Fail            & 0.5023      & 0.4219       & 1172.2      & 1275.4      \\
        Win95PTS        & 76                    & 0.4856      &   Fail           & 932.6       &  Fail            & 0.4912      & 0.3250        & 912.4       & 994.07      \\ \bottomrule
    \end{tabular}}
\end{table*}

\section{Experimental Evaluation for LLM-guided Causal Discovery}
\label{app_exsec3_5}

We conducted experiments to evaluate the effectiveness of LLMs as initialization tools and evolutionary operators in causal learning, examining their impact on algorithmic performance. We utilized 10 real-world causal network datasets from the causal community benchmark bnlearn, which offer well-defined variables and reliable prior knowledge. These datasets vary in scale: small (e.g., Asia, Sachs), medium (e.g., Insurance, Water, Alarm, Barley), large (e.g., Hepar II, Win95PTS), and extra-large (e.g., Pathfinder, Andes). Each dataset contains 1000 samples, and each experiment was independently run 30 times to compute the mean results for robustness.

\textbf{Initial Population Initialization}: We compared LLM-based initialization with two classical methods, i.e., Conditional independence-based method (CI-Based), which removes independent edges based on conditional independence tests to reduce the search space, and Mutual information-based method (MI-Based), which prunes the initial search space using mutual information as a relevance measure. We employed the top three performing LLMs from Section \ref{sec3} (GPT-4, Claude-3, GPT-3.5) to assist in search space reduction. Variable information and background knowledge were formatted into specific prompts and input into the LLM to obtain relationship scores. 

\begin{figure}[t]
  \centering
  \includegraphics[width=12cm]{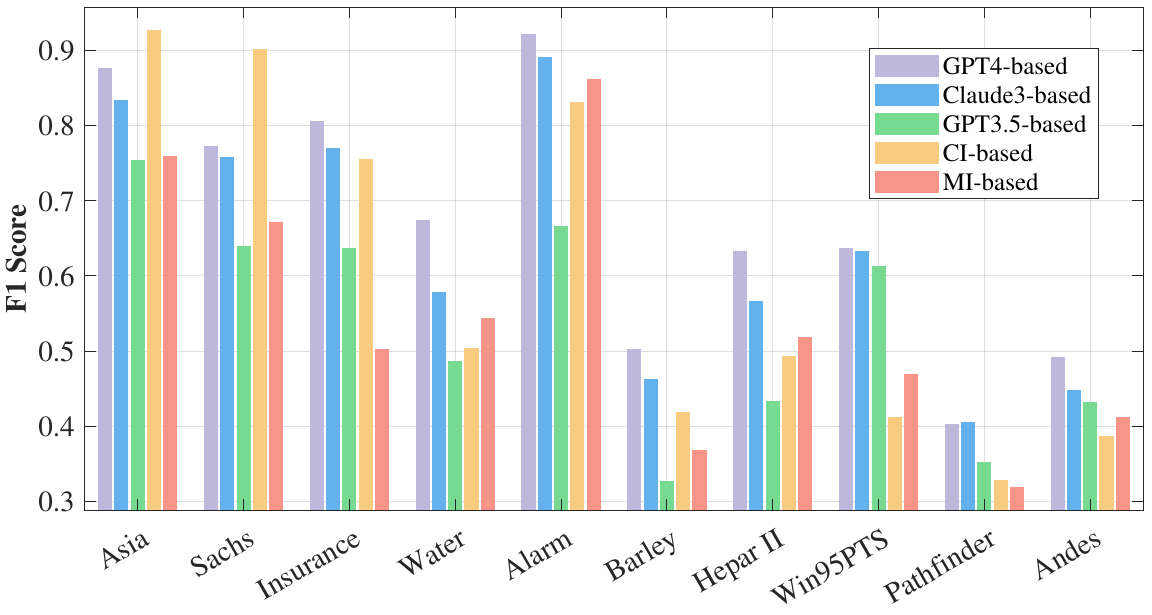}
  \caption{The F1 Score of different initialization methods on 10 bnlearn datasets.}\label{app_ex_6_1}
\end{figure}

Figure \ref{app_ex_6_1} presents the experimental results. On small datasets, the CI-based method performs best due to its accurate reduction of the search space. However, on medium and large datasets, LLM-based search space reduction significantly outperforms other strategies. In some datasets, LLM-based performance is comparable to a combined CI-MI method, likely due to complex variable names limiting LLMs' effectiveness. This indicates that LLMs exhibit substantial potential in handling large-scale datasets by leveraging extensive prior knowledge to eliminate irrelevant variables and reduce search complexity, thereby providing a more efficient starting point for causal learning.

\textbf{LLM-based Evolutionary Operators}: In the evolutionary operation phase, we compare LLM-assisted evolution with traditional evolutionary methods. Specifically, we consider two traditional approaches: (1) EO1, where the crossover operation employs a conventional uniform crossover, and mutation is performed using standard bit-flipping; (2) EO2, where the crossover stage adopts a traditional parent-based crossover, while mutation remains standard bit-flipping. For the LLM-assisted evolutionary process, we adopt the Tree of Thoughts (ToT) prompting strategy, which provides the LLM with optimization task information, contextual examples, and expected output formats to guide crossover and mutation operations. Selection is performed using tournament selection. In contrast, EO1 and EO2 follow their respective predefined evolutionary operations.

\begin{figure}[t]
  \centering
  \includegraphics[width=3.3cm]{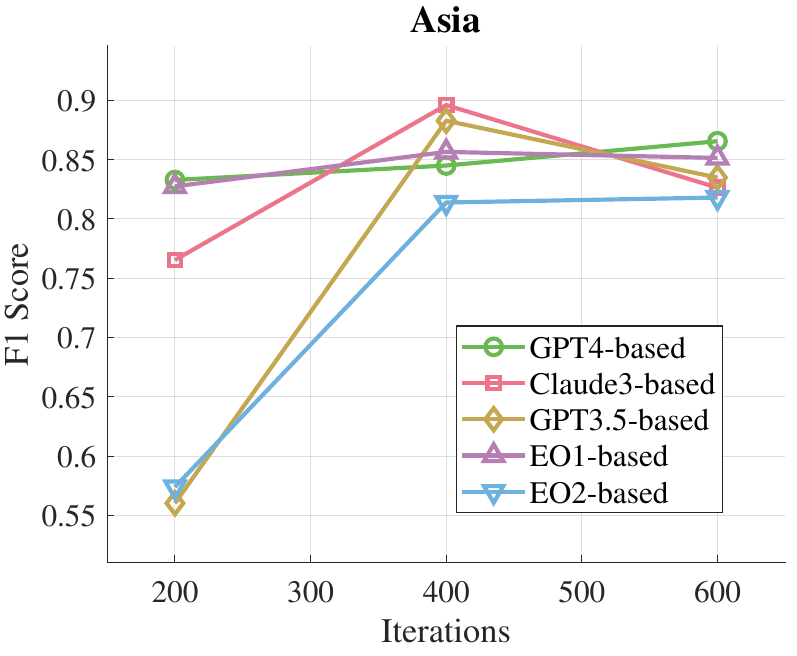}
  \includegraphics[width=3.3cm]{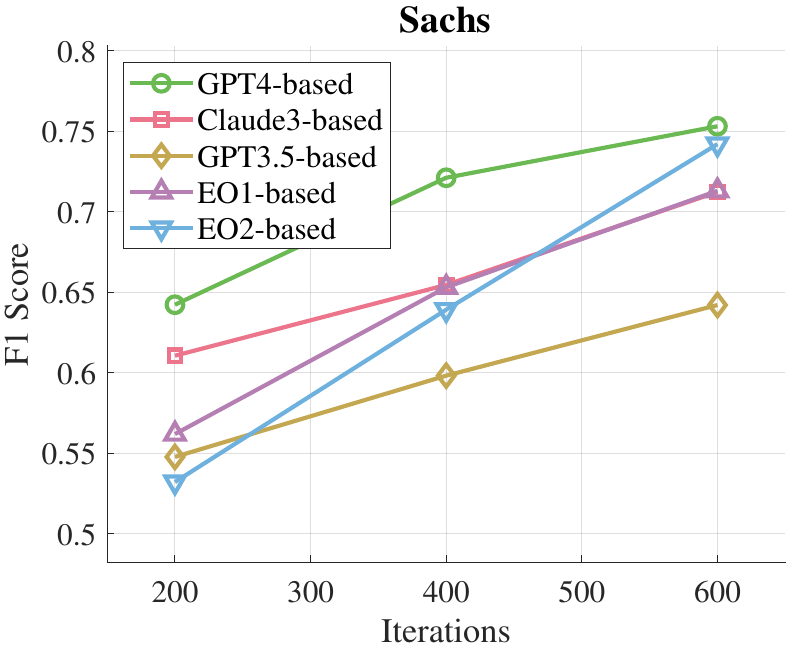}
  \includegraphics[width=3.3cm]{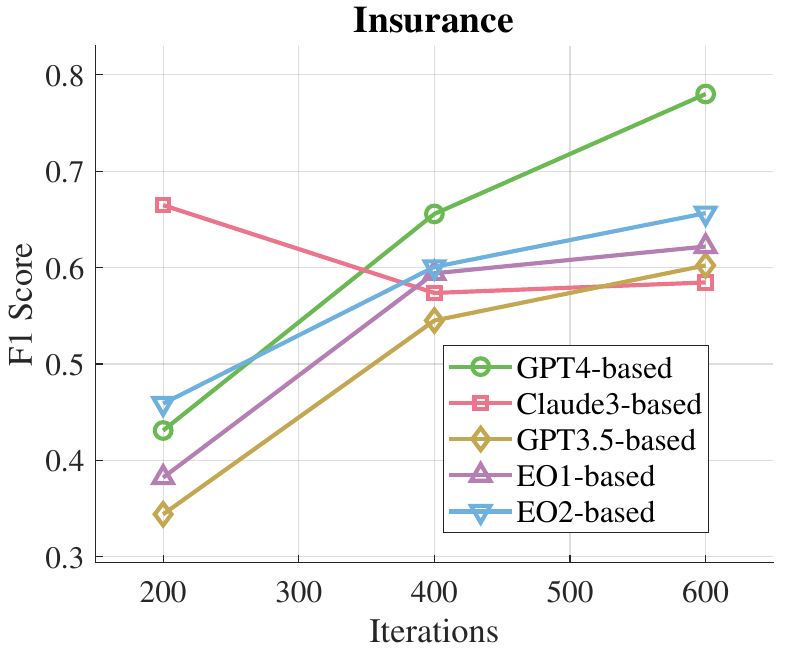}
  \includegraphics[width=3.3cm]{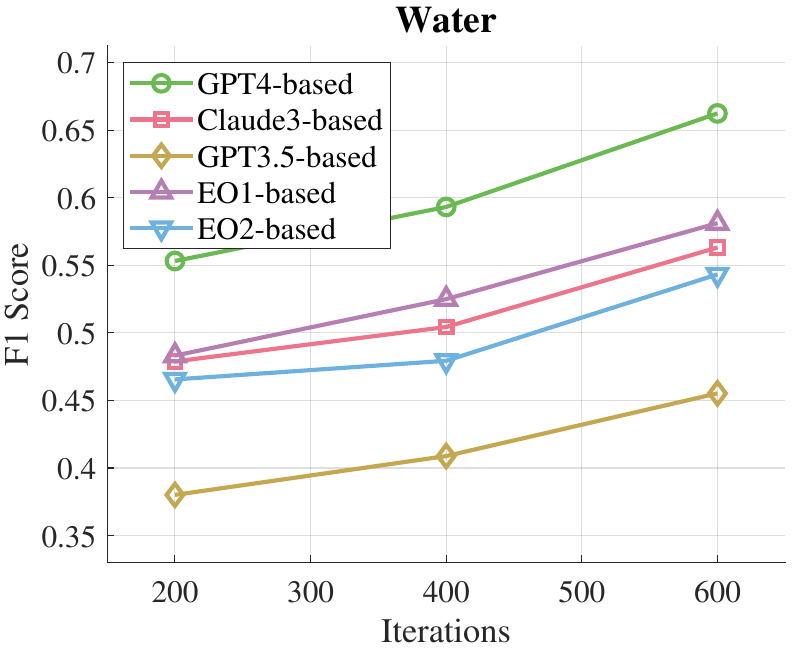}
  \includegraphics[width=3.3cm]{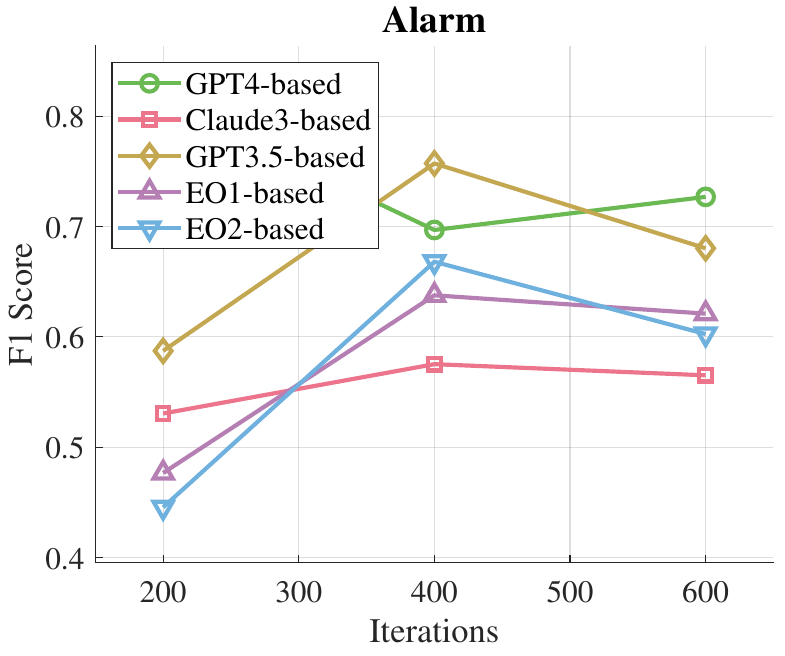}
  \includegraphics[width=3.3cm]{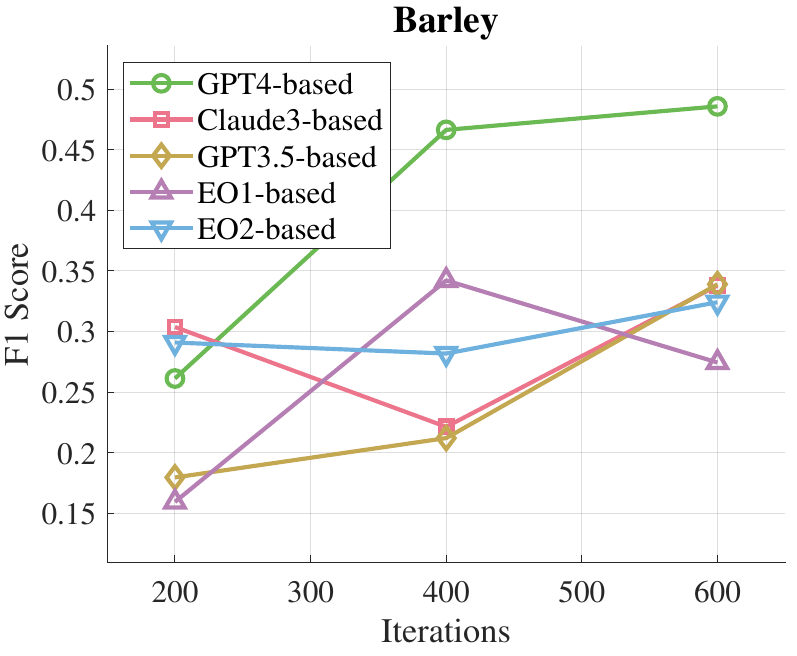}
  \includegraphics[width=3.3cm]{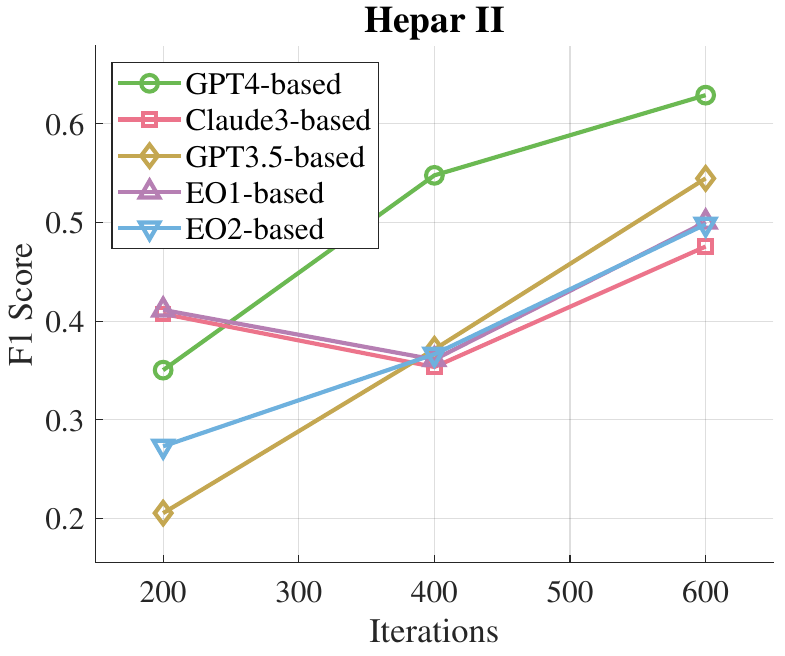}
  \includegraphics[width=3.3cm]{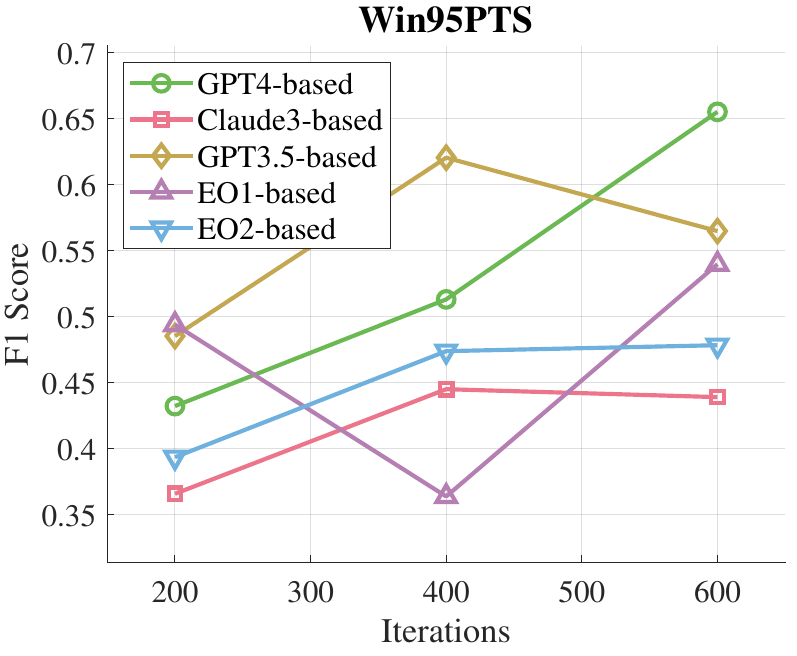}
  \includegraphics[width=3.3cm]{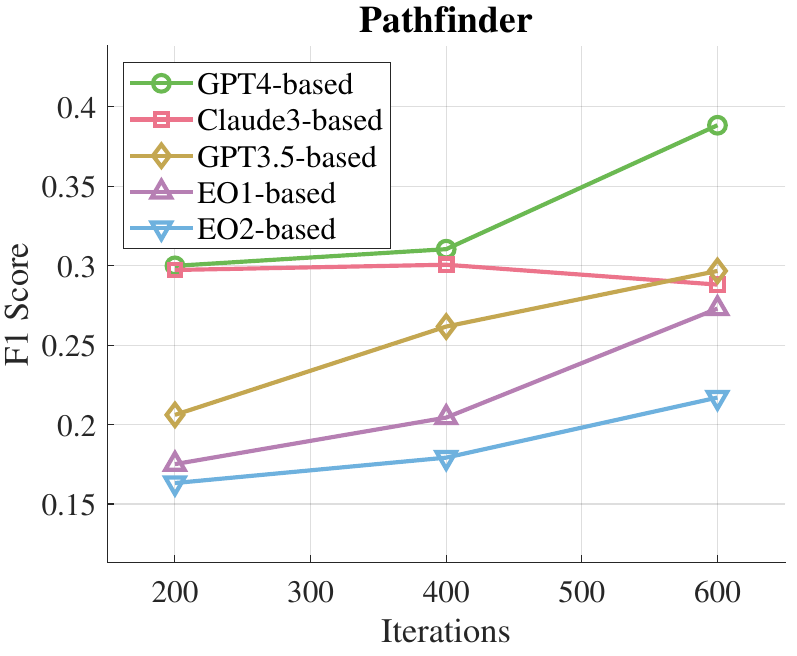}
  \includegraphics[width=3.3cm]{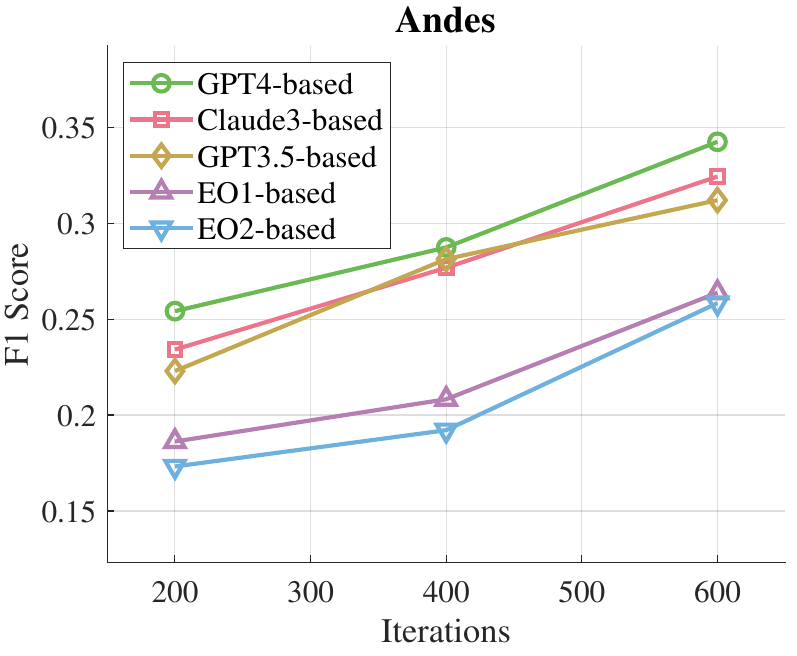}
  \caption{The F1 Score of different evolutionary operators on 10 bnlearn datasets.}\label{app_ex_6_2}
\end{figure}

To evaluate the effectiveness of these approaches, we conduct experiments under different numbers of fitness evaluations $(200,400,600)$, observing the convergence behavior of each method. As illustrated in Figure \ref{app_ex_6_2}, among all LLM-based methods, GPT-4 exhibits the best performance across different fitness evaluation budgets. This advantage can be attributed to its larger training corpus, broader background knowledge, and stronger reasoning capabilities. Although the performance of GPT-3.5 and Claude 3 is slightly inferior to that of GPT-4, they achieve results comparable to the traditional evolutionary operation methods EO1 and EO2. This validates the effectiveness of LLM-assisted evolutionary operations, demonstrating that well-trained LLMs, with their extensive pretraining on world knowledge, can match or even surpass traditional evolutionary crossover and mutation strategies. These findings suggest that incorporating LLMs as evolutionary operators can enhance the search efficiency and accuracy of causal learning algorithms, enabling faster discovery of optimal causal structures.

\textbf{Potential Extensions of LLM-Assisted Evolutionary Optimization}: Beyond the aforementioned approaches, LLMs offer several additional possibilities for assisting causal structure optimization. One promising direction is the dynamic adjustment of search strategies. As the search progresses, the LLM can be utilized to dynamically assess the search state and adjust the strategy accordingly. If the LLM detects that the search algorithm has stagnated in a particular region—i.e., multiple consecutive iterations yield no significant improvement—it can leverage its understanding of the current search space to suggest modifying the step size or adjusting evolutionary parameters such as mutation probability and crossover rate. These adjustments enable the search algorithm to escape local optima and continue exploring more promising solutions. Furthermore, the LLM can utilize accumulated search information to predict regions with a higher likelihood of containing optimal solutions. By guiding the search algorithm toward these regions, the overall search efficiency can be further enhanced. This capability highlights the potential of LLMs not only as passive evolutionary operators but also as active agents in shaping adaptive and intelligent search strategies.

\section{Empirical Studies for Alternative Views}
\label{app_exsec3_6}

\textbf{Manipulability of Experimental Results}: To validate our argument, we conducted a series of experiments on soft-constrained scoring functions using GPT-4. Specifically, we selected 14 small datasets from `bnlearn` and tested two types of prompts on each dataset. The first type, high-quality background knowledge prompts, underwent rigorous manual curation, thorough verification, and careful refinement to eliminate redundant information. These prompts contained direct causal statements and ensured close textual proximity between variables, often explicitly stating causal relationships. The second type, general background knowledge prompts, sourced information from online repositories such as Wikipedia, contained substantial redundant information, and lacked guarantees regarding the correctness of causal knowledge, as illustrated in Figure \ref{app_ex_6_1}.

\begin{figure}[t]
  \centering
  \includegraphics[width=14cm]{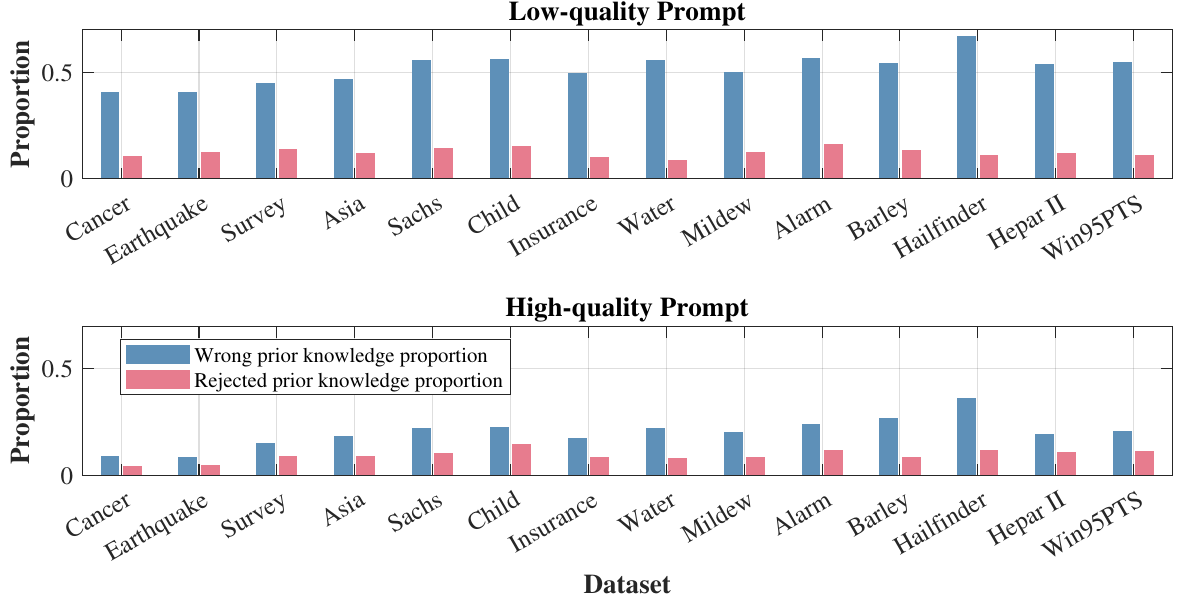}
  \caption{The wrong prior knowledge proportion and rejected prior knowledge proportion on different datasets under low-quality and high-quality prompts.}\label{app_ex_5}
\end{figure}

Experimental results clearly demonstrate that, despite substantial differences in the proportion of incorrect prior knowledge under the two prompt settings, the proportion of incorrect prior knowledge rejected by the soft-constrained scoring function remains consistently low. This finding indicates that soft-constrained scoring functions are highly limited in their ability to filter out incorrect priors. Under low-quality prompts, most erroneous priors generated by LLMs are retained. However, by improving prompt quality, the likelihood of LLMs generating incorrect priors decreases, consequently reducing the proportion of errors undetected by the soft constraints. Importantly, these reductions in errors do not stem from the capability of the soft constraints themselves but rather from the manual enhancement of textual quality at an additional cost.

\end{document}